\documentclass[journal]{IEEEtran}

\ifCLASSINFOpdf

\else

\fi

\usepackage{color}
\usepackage{amssymb}
\usepackage[colorlinks=False,linkcolor=blue]{hyperref}
\usepackage{amsmath}
\usepackage{graphicx}
\usepackage{caption}
\usepackage[dvipsnames]{xcolor}
\usepackage{subcaption}
\graphicspath{ {images/} }
\usepackage{float}
\usepackage{amsthm}
\usepackage{listings}
\usepackage{algorithm,algcompatible}
\makeatletter
\def\munderbar#1{\underline{\sbox\tw@{$#1$}\dp\tw@\z@\box\tw@}}
\makeatother

\usepackage{algorithm}
\usepackage{algorithmicx}
\usepackage{algpseudocode}
\usepackage{makecell}
\usepackage{booktabs}       
\usepackage{url}
\usepackage{multirow}
\usepackage{lipsum}
\usepackage{bbm}
\usepackage{soul}
\usepackage{multirow}
\usepackage{setspace}
\usepackage{stmaryrd}
\usepackage{color}
\usepackage{xcolor}
\usepackage{nomencl}
\makenomenclature

\newtheorem{remark}{Remark}

\newtheorem{definition}{Definition}

\newcommand{\R}{\mathbb{R}}
\newcommand{\Z}{\mathbb{Z}}
\newcommand{\x}{\mathbf{x}}

\newcommand{\s}{\mathbf{s}}

\usepackage{CJKutf8}

\begin{document}

\title{A Diversity Analysis of Safety Metrics Comparing Vehicle Performance in the \\ Lead-Vehicle Interaction Regime}

\author{Harnarayan Singh$^{1}$, Bowen Weng$^{1}$, Sughosh J. Rao$^{1}$, Devin Elsasser$^{2}$% <-this % stops a space

\thanks{This work was supported by National Highway Traffic Safety Administration (“NHTSA”) Contract 693JJ918D000019 Task Order 693JJ921F000036. \textit{(Corresponding author: Bowen Weng.)}}%
\thanks{$^{1}$Harnarayan Singh, Bowen Weng, and Sughosh J. Rao are with the Transportation Research Center Inc., East Liberty, Ohio, USA on assignment to National Highway Traffic Safety Administration, USA (e-mail: harnarayan.singh.ctr@dot.gov; bowen.weng.ctr@dot.gov; sughosh.rao.ctr@dot.gov).}% 
\thanks{$^{2}$Devin Elsasser is with the National Highway Traffic Safety Administration, USA (e-mail: devin.elsasser@dot.gov).}%
\thanks{A modified manuscript of this preprint has been accepted to be published as a regular paper at IEEE Transactions on Intelligent Transportation Systems.}
}

\maketitle

\begin{abstract}
Vehicle performance metrics analyze data sets consisting of subject vehicle’s interactions with other road users in a nominal driving environment and provide certain performance measures as outputs. To the best of the authors' knowledge, the vehicle safety performance metrics research dates back to at least 1967. To date, there still does not exist a community-wide accepted metric or a set of metrics for vehicle safety performance assessment and justification. This issue gets further amplified with the evolving interest in Advanced Driver Assistance Systems and Automated Driving Systems. In this paper, the authors seek to perform a unified study that facilitates an improved community-wide understanding of vehicle performance metrics using the lead-vehicle interaction operational design domain as a common means of performance comparison. In particular, the authors study the diversity (including constructive formulation discrepancies and empirical performance differences) among 33 base metrics with up to 51 metric variants (with different choices of hyper-parameters) in the existing literature, published between 1967 and 2022. Two data sets are adopted for the empirical performance diversity analysis, including vehicle trajectories from normal highway driving environment and relatively high-risk incidents with collisions and near-miss cases. The analysis further implies that (i) the conceptual acceptance of a safety metric proposal can be problematic if the assumptions, conditions, and types of outcome assurance are not justified properly, and (ii) the empirical performance justification of an acceptable metric can also be problematic as a dominant consensus is not observed among metrics empirically.
\end{abstract}


\IEEEpeerreviewmaketitle

\section{Introduction}\label{sec:intro}
\begin{figure}
    \centering
    \includegraphics[width=.48\textwidth]{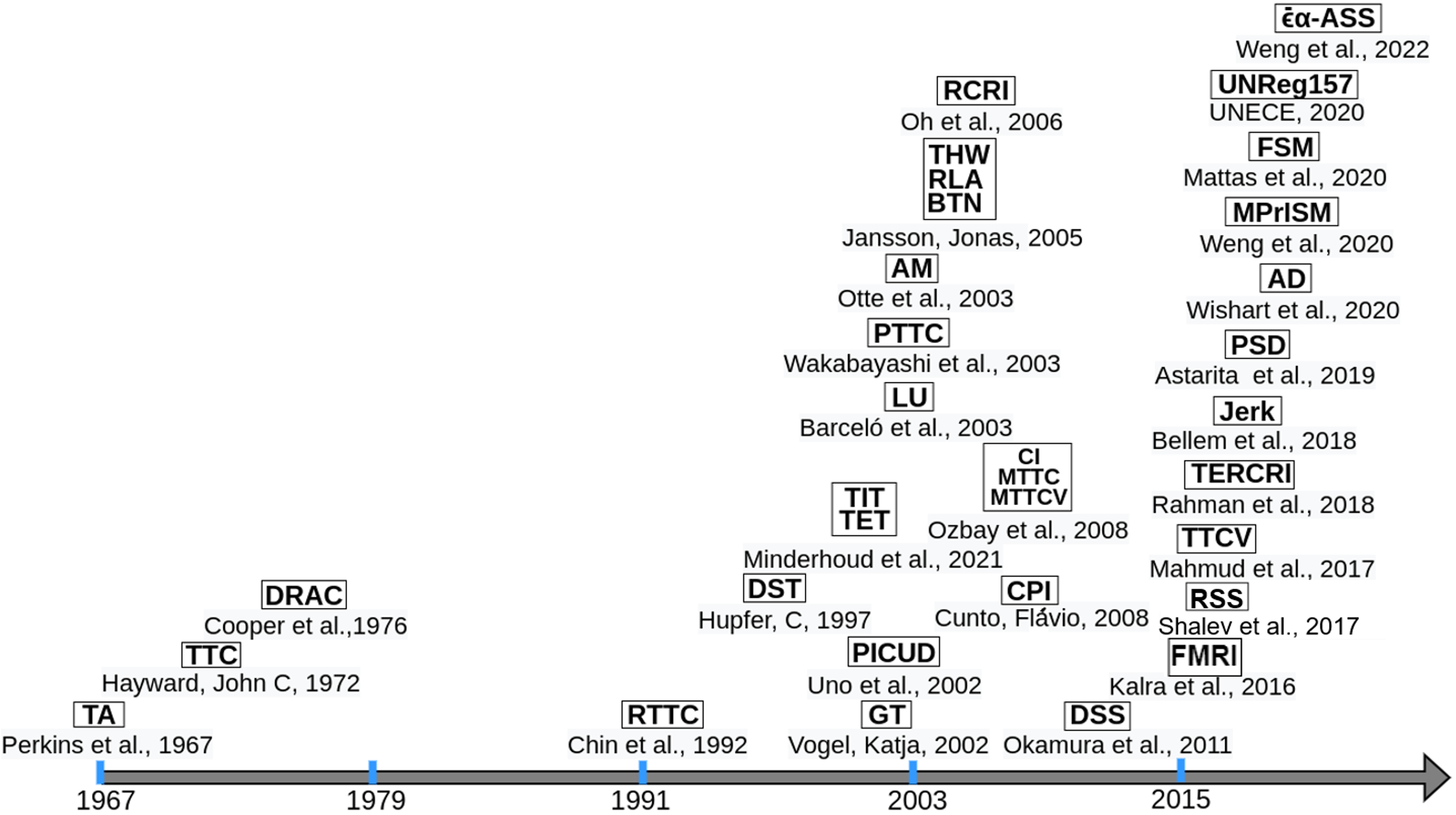}
    \caption{Some of the metrics studied in this paper along with their first-time appearance in the literature. Metrics are illustrated by the same acronyms defined later in TABLE~\ref{tab:metrics}.}
    \label{fig:overview}
    \vspace{-4mm}
\end{figure}
In this paper, a vehicle performance metric is considered as a mapping with a certain input and an output. The input is a data set that characterizes the interactive motion between the subject vehicle (SV) (or a group of SVs) and other road users, along with some environmental conditions, within a specific Operational Design Domain (ODD). The output is a performance measure with various physical properties and mathematical notions of interpreting the performance of the SV (or SVs) interacting with other road users in the given ODD, revealed by the extracted data set. This particular formulation of metrics is primarily intended for driving performance evaluation and assessment, performance benchmark, and standardization use cases, where a data set is presented as it stands. This is different from the run-time verification applications~\cite{lin2023commonroad, shalev2017formal} where the safety performance is justified online to facilitate the planning and control of vehicles. Note some metrics may be used for both post-processing evaluation and run-time verification purposes. Throughout the remainder of this paper, one considers the specific ODD of the lead-vehicle interaction on straight-road segments. This is chosen to limit maneuver based variables and focus on comparison of metrics. The lead-vehicle interaction ODD (the formal definition is explained in detail in Section~\ref{sec:preliminaries:lead-vehicle}) is also one of the most commonly studied ODDs in the vehicle safety literature and is compatible with all metrics studied in this paper~\cite{zhao2018accelerated, fan2018data, weng2023finite}. Extensions can be made to other ODDs and other metrics using the proposed diversity analysis framework.

One of the first vehicle safety performance metrics was proposed in 1967, referred to as the Time to Accident~\cite{perkins1967traffic}. To date, the world has experienced two peak periods with tremendous amount of new metric proposals and is currently going through the third peak, as illustrated in Fig.~\ref{fig:overview}. The first peak back in the 1960s coincided with the birth of vehicle safety regulations. Many classic safety performance metrics were created, including the Time-to-Collision (TTC) metric~\cite{hayward1972near}, which is one of the most well-adopted safety performance measurements in the study of rear-end collision avoidance. The second peak started around the 1990s and lasted for almost two decades. It was accompanied by some of the most important pioneering works in the intelligent vehicle field, such as the DARPA Grand Challenge~\cite{seetharaman2006unmanned} in 2004. Recently, with the emerging interest in Advanced Driver Assistance Systems (ADAS) and Automated Driving Systems (ADS), the work domain of safety metrics has been extended to traffic interactions involving a broader group of road users, such as multi-vehicles and pedestrians. However, after over half a century of research efforts, the community has still not reached a common consensus on an acceptable metric (or a set of metrics) for performance evaluation of vehicles, especially ADAS or ADS equipped vehicles.

\begin{table}[!b]
    \caption{An overview of the 33 base metrics (the superscript~* in the third column denotes the acronym created by the authors, as one is not available in the original publication).}
    \label{tab:metrics}
    % \vspace{2mm}
    \centering
    \resizebox{0.48\textwidth}{!}{%
    \begin{tabular}{l|l|c}
    \hline
    Index & Metric Name   & Acronym \\ \hline
    1 & Time to Collision~\cite{hayward1972near} & TTC \\
    2 & Potential Time to Collision~\cite{wakabayashi2003traffic} & PTTC \\
    3 & Modified Time to Collision~\cite{Ozbay2008Derivation} & MTTC   \\
    4 & Model Predictive Instantaneous Safety Metric~\cite{weng2020model} & MPrISM  \\      
    5 & Potential Index for Collision with Urgent Deceleration~\cite{uno2002microscopic} & PICUD   \\
    6 & Difference of Space-distance and Stopping-distance~\cite{okamura2011impact}  & DSS   \\
    7 & Deceleration Rate to Avoid the Crash~\cite{cooper1976traffic} & DRAC   \\
    8 & Deceleration to Safety Time~\cite{hupfer1997deceleration} & DST   \\
    9 & Required Longitudinal Acceleration~\cite{Jansson2005CollisionAT} & RLA   \\
    10 & Reciprocal of TTC~\cite{chin1992quantitative} & RTTC   \\    
    11 & Brake Threat Number~\cite{Jansson2005CollisionAT} & BTN   \\
    12 & Time to Accident~\cite{perkins1967traffic} & TA   \\ 
    13 & Crash Potential Index~\cite{cunto2008assessing} & CPI \\
    14 & Time Exposed TTC~\cite{minderhoud2001extended} & TET   \\
    15 & Time Integrated TTC~\cite{minderhoud2001extended} & TIT   \\
    16 & Rear-end Collision Risk Index~\cite{oh2006method} & RCRI   \\
    17 & Time Exposed Rear-end Collision Risk Index~\cite{rahman2018longitudinal} & TERCRI   \\
    18 & Time to Collision Violation~\cite{mahmud2017application} & TTCV   \\ 
    19 & Modified Time to Collision Violation~\cite{Ozbay2008Derivation} & MTTCV   \\ 
    20 & Responsibility Sensitive Safety~\cite{shalev2017formal} & RSS  \\ 
    21 & Fuzzy Safety Model~\cite{mattas2020fuzzy} &  FSM  \\
    22 & UN Regulation 157~\cite{united2020proposal}  & UNReg157$^*$ \\
    23 & Crash Index~\cite{Ozbay2008Derivation} & CI   \\
    24 & Proportion of Stopping Distance~\cite{astarita2019traffic} & PSD   \\ 
    25 & Aggressive Driving~\cite{wishart2020driving} & AD  \\
    26 & Accident Metric~\cite{otte2003scientific} & AM   \\ 
    27 & Jerk~\cite{bellem2018comfort} & Jerk$^*$ \\
    28 & Gap Time~\cite{vogel2002characterizes} & GT \\
    29 & Time Headway~\cite{Jansson2005CollisionAT} & THW   \\
    30 & Level of Unsafety~\cite{barcelo2003safety} & LU$^*$   \\ 
    31 & Collision Rate & CR$^*$ \\
    32 & Failure-free Miles Risk Inference~\cite{kalra2016driving} & FMRI$^*$ \\
    33 & $\bar{\epsilon}\alpha$-Almost Safe Set~\cite{weng2023finite} &$\bar{\epsilon}\alpha$-ASS$^*$ \\ \hline
    \end{tabular}%
    }
    % \vspace{-5mm}
\end{table}

In general, a proposed vehicle safety performance metric argues its acceptability through \emph{constructive proof} and/or \emph{empirical evidence}. Metrics with constructive proof come with conditions and assumptions under which the safety performance outcome is justified analytically. On the other hand, empirical evidence demonstrates the performance of a proposed metric by showing numerical outcomes that align with intuitions and common expectations. To help bridge the gaps among various metrics with different constructive proofs and empirical outcomes in their respective original proposals, this paper seeks to, formally and empirically, understand the discrepancies and correlations among some of the well-adopted metrics in the literature. This is referred to as the \emph{diversity analysis}.

From the metric construction perspective, the diversity analysis is challenging, as metrics are proposed for different working purposes, with different assumptions and logical reasoning. The constructive diversity analysis is expected to capture the formulation discrepancies and to classify metrics into a certain finite number of categories by the constructive nature. Some recent surveys~\cite{elli2021evaluation,jammula2022evaluation} distinguish metrics by the output units (e.g., distance-based metrics and time-based metrics). Such an intuitive classification criterion fails to capture the fundamental constructive property of a metric and is not compatible with the broader spectrum of metrics studied in this paper (e.g., CI (defined later in Appendix~\ref{apx:constructive_da:overview}) maps to an output with unit $\mathrm{m^2/s^3}$). The idea of microscopic and macroscopic risk metrics~\cite{junietz2019microscopic}, where the metrics applicable to individual traffic scene like TTC are classified as microscopic and overall average measure related metrics like collision rate are classified as macroscopic, is also limited as it fails to reveal the constructive insights of metrics. Griffor et al.~\cite{griffor2019workshop} and some other works have been using the leading \& lagging measure characterizations. This captures some, but not all, of the constructive discrepancies. In particular, the lagging measure category fails to differentiate between metrics with observed outcomes and those with statistically inferred outcomes. One can refer to Section~\ref{sec:constructive_da} later for a detailed discussion on this issue.

Moreover, the diversity analysis gets more challenging with the empirical performance study as (i) metric outputs are of various physical properties and data types that are not directly comparable (e.g., comparing distance with time, comparing a Real scalar with a Boolean value output), and (ii) metric outputs of the same physical property and data type are not necessarily of the same admissible set of values (e.g., both TTC and MPrISM take the notion of ``time" as the output, but have different admissible ranges of values). Within the empirical performance comparison topic, existing works~\cite{kidambi2022sensitivity,mattas2022driver} either identify if metrics agree with each other on the justification of a given traffic scene being safe or unsafe, or compare metric outcomes of the same physical meanings and admissible values, such as the TTC variants in~\cite{kidambi2022sensitivity}. This has significantly restricted their study to a sub-set of metrics considered in this paper.

To resolve the aforementioned challenges, the main contributions of this paper are further summarized as follows.

\subsection{Main Contributions}
\textbf{A Formal Constructive Diversity Analysis of Metrics}: This paper covers 33 base metrics for vehicle safety performance justification published in the literature between 1967 and 2022. Through an analytical study of metric formulations, the authors further propose a framework that classifies metrics into three categories, along with other simple extensions and combinations that cover all 33 studied metrics, as well as some of the unmentioned metrics in the literature. This framework helps the understanding of fundamental discrepancies and correlations among metrics. It can be further extended to other metrics mentioned in Section~\ref{sec:oliterature} that are not part of the 33 base metrics. Finally, the framework forms a ``recipe" of metric creation and can be re-directed to formulate metrics that do not necessarily exist in the literature. 

\textbf{An Empirical Performance Diversity Analysis of Metrics}: Provided with the same group of extracted data sets as the input, the empirical output discrepancies among a total number of 51 metric variants (due to different choices of hyper-parameters) are studied. To compare the empirical discrepancies among metrics of different output formats, the authors further take advantage of the fact that every metric output is either monotonic or consists of elements that are monotonic w.r.t. a certain notion of performance justification (e.g., large and small TTC values imply low and high risk, respectively)~\cite{hayward1972near, weng2020model, uno2002microscopic, okamura2011impact, cooper1976traffic, chin1992quantitative, Jansson2005CollisionAT, cunto2008assessing, oh2006method, barcelo2003safety, kalra2016driving}. As a result, this paper presents a two-step framework for empirical performance diversity analysis of different metrics, including (i) transferring the metric outcome to a classification categorization by taking advantage of the aforementioned property, and (ii) applying standard justification methods that compare classification outcomes, such as the \emph{agreement index} (AID) and the \emph{precision recall study}~\cite{saito2015precision}. Both methods come with the value of zero and one indicating complete disagreement and complete agreement, respectively. The pairwise agreement among all studied metrics among all extracted data sets obtains an average AID of 0.650 (with a standard deviation of $\pm 0.262$). That is, the studied metrics empirically disagree with each other heavily without a dominant consensus. This further implies that \emph{the commonly adopted practice of an empirical justification of the acceptable metric may be problematic.}

Note that this study is of survey value given the broad coverage of state-of-the-art metrics in various types. But more importantly, this paper is beyond the typical scope of a survey that enumerates metrics and analyzes each one individually. The diversity analysis provides a unified view of metrics analytically and numerically. The proposed methodology can be further extended to other metrics and other ODDs. Moreover, it is not the direct intention of this paper to propose/recommend a metric or a certain set of metrics for vehicle safety performance assessment and justification. This paper is primarily concerned with the diversity analysis w.r.t. the two aforementioned perspectives, along with a hope to encourage the community-wide metric acceptance in the future.

\subsection{Construction}
The overall construction of the paper is as follows. Section~\ref{sec:preliminaries} reviews the terminologies used throughout the paper. Section~\ref{sec:constructive_da} studies the constructive discrepancies among the 33 base metrics. Section~\ref{sec:performance_da} further extends the diversity study to the empirical performance discrepancies. Finally, conclusions and future work are discussed in Section~\ref{sec:conclusion}.

\textbf{Notation: } The set of Real and positive Real numbers are denoted by $\R$ and $\R_{>0}$, respectively. $\Z_{>0}$ denotes the set of all positive integers and $\Z_N=\{1,\ldots,N\}$. $|X|$ is the cardinality of the set $X$, e.g., for a finite set $D$, $|D|$ denotes the total number of points in $D$. $|\x|$ can also denote the absolute value for some $\x \in \R^n$. $\text{sign}(x)$ is the sign function that returns a value in $\{-1,0,1\}$. $\llbracket i, j \rrbracket$ denotes the Kronecker delta function as 
\begin{equation}\label{eq:kdf}
    \llbracket i, j \rrbracket = 
    \begin{cases}
    1 & \text{if } $i = j$ \\
    0 & \text{otherwise}
  \end{cases}.
\end{equation}
Some commonly used acronyms are also adopted including i.i.d. (independent and identically distributed), w.r.t. (with respect to), and w.l.o.g. (without loss of generality). 

\section{Preliminaries}\label{sec:preliminaries}
This section formally reviews the terminologies associated with the construction of an ADS performance metric, including its input data set and output performance measures.

\subsection{Terminologies}\label{sec:preliminaries:terminologies}
Let $\mathcal{S} \subseteq \R^n$ denote the set of observed states associated with the interactive motion between a certain SV, other road users and driving environment, where $n \in \Z$ denotes the state dimension. Examples of the state $\s \in \mathcal{S}$ include global positions and velocities of all vehicles. One can also include non-dynamics related properties such as vehicle color, weather condition, and road curvature, to name a few. Note that one and only one SV is allowed in each $\s \in \mathcal{S}$.

Each subset of $\mathcal{S}$, i.e., $O \subseteq \mathcal{S}$, forms what one commonly refers to as the ODD. Intuitively, the ODD is the operating conditions under which the SV is expected to achieve a certain functionality. Such operating conditions are formally characterized by the set of admissible states in $O$. It is thus immediate that the construction of an ODD is not unique and each SV can have multiple ODDs. One can refer to~\cite{czarnecki2018operational,weng2023finite} for various ODD examples within the ADS context. 

As the traffic interactions evolve with time within a given ODD or a combined set of ODDs, a data set is thus established as follows.
\begin{definition}
A data set, $\mathcal{D} \subset \mathcal{S} \times T$, is a set of time-dependent states collected through a certain testing data acquisition system (DAS) of a constant frequency $f$. W.l.o.g., $T$ is a set of positive integers representing the set of time steps.
\end{definition}
Note all $\s(t) \in \mathcal{D}$ (i.e., $\s(t) \in \mathcal{S}$ at the given time $t \in T$) do not necessarily share the exact same SV. For example, the HighD data set~\cite{krajewski2018highd} (specifically the combined HighD and HighD Plus data set) we introduce later, consists of a mixture of cars and trucks driven by different human drivers on German highways, any of which can be considered the SV. Within each data set, the SV (or SVs) should exhibit a statistically consistent and rational behavior. The naturalistic human driver behavior as well as those integrated with ADAS or ADS typically all satisfy this condition. Moreover, this paper primarily focuses on SV interaction with vehicles, referred to as Principal Other Vehicles (POVs). Other road users, such as pedestrians, are out of scope for this study.

Occasionally, a data set can also be viewed as a combined set of \emph{incidents}, with the incident being defined as follows.
\begin{definition}
An incident, $\mathcal{I}=\{\s(t)\}_{t \in {i+1, \dots, i+\tau}}$, is a set of states consecutive in time sharing the same SV. The number of time steps in a given $\mathcal{I}$, $\tau$, is referred to as the \emph{length} of the incident.
\end{definition}
A data set $\mathcal{D}$ thus contains multiple incidents of various lengths. Note that similar terms are also used in the literature as alternatives to the defined incident, such as scenario~\cite{ulbrich2015defining}, testing case~\cite{thorn2018framework}, event, and episode.

\begin{remark}
For the remainder of this paper, the notion of a SV can be a specific subject vehicle (e.g., a 2000 Ford Fusion) or a group of SVs sharing the same type of driving behavior (e.g., human driven cars). The authors do not differentiate the notion of a single SV from a group of SVs that share similar nature of concern. One can refer to Section~\ref{sec:performance_da} for specific related examples.
\end{remark}

Note that the collected raw data set can be quite large and can cover a variety of interactions between the SV and the POVs in different ODDs. For safety analysis purposes, one typically interprets the study of a given data set in groups, with each group confined to a particular ODD centered around a particular SV. This creates an \emph{extracted} subset of $\mathcal{D}$ as 
\begin{equation}\label{eq:ext_data}
    \mathcal{D}_{O,k} \subseteq {\mathcal{D}}.   
\end{equation}
that denotes the set of states from the data set $\mathcal{D}$, occurred within the given ODD $O$, and associated with the selected SV represented by a certain index $k$. Moreover, to ensure the fairness for performance benchmark and comparison purposes, the data sets are expected to be collected with the same testing strategy for all SVs. Although the rigorous definition of the similar testing strategy can be obscure~\cite{weng2023comparability}, it generally indicates to collect data sets within the same naturalistic driving environment~\cite{feng2023dense}, against the same set of testing scenarios~\cite{euro2017test}, or following the same distribution of testing cases~\cite{feng2020testing}, as shown in Section~\ref{sec:performance_da}.

\subsection{The Lead-Vehicle Interaction ODD}\label{sec:preliminaries:lead-vehicle}
Throughout this paper, the lead-vehicle interaction domain, $O_l$, is selected as the primary ODD of interest. The SV oriented ODD, $O_0$, is also introduced as it is a sub-set of $O_l$. Details are presented as follows.

\textbf{The SV Oriented ODD}: The set $O_0$ consists of dynamic states that are strictly associated with the SV (e.g., the SV's position, velocity, acceleration, and lane index) and other environmental conditions that directly affect the SV (e.g., weather condition, road friction, and lane curvature). The particular feature selection and specifications are subject to the raw data set $\mathcal{D}$ in general. Given the two data sets analyzed by this paper, $O_0$ is further confined to a single-lane straight road segment. In practice, constant conditions are omitted in the ODD notation, hence the particular $O_0$ studied by this paper admits features such as SV position, velocity, and acceleration.

\textbf{The Lead-Vehicle Interaction ODD}: The lead-vehicle interaction ODD, $O_l$, expands $O_0$ (i.e., $O_0 \subset O_l$) by adding to the domain a lead POV, sharing the same lane with the SV. Given the straight-road condition mentioned above, we have $O_l$ as the state space concerning position, speed, acceleration, and heading angle of the follower SV and the leading POV. Moreover, the distance headway (DHW) between the two vehicles is formally specified as the $\ell_2$-norm distance between the center points of the SV's front bumper and the lead POV's rear bumper. In practice, the design of $O_l$ also incorporates other selected features of the SV and the leading POV, which include length, width, etc. Note that the vehicle driving backward is not part of the ODD, and the leading POV stays in front of the follower SV persistently. In the remainder of this paper, $O_l$ and $O$ are interchangeable, as $O_l$ is the only ODD of interest for this study.

The ODD design admits features and conditions that are available from the raw data set and are deemed important for ADS performance analysis. Some features are not included in the design of $O$ or are not even part of $\mathcal{D}$. They are thus considered as disturbances and uncertainties (e.g., weather condition and road gradient), mostly following a certain unknown but fixed distribution embedded in the construction of the raw data set. 

Recall the extracted data set notion of \eqref{eq:ext_data}, we have $\mathcal{D}_{O_0,k}$ and $\mathcal{D}_{O_l, k}$. Note that metrics compatible with $\mathcal{D}_{O_0,k}$ are also compatible with $\mathcal{D}_{O_l, k}$, as $O_0 \subset O_l$. This forms the input of an ADS performance metric as introduced in the following subsection.

\subsection{The ADS Performance Metric}\label{sec:preliminaries:metrics}
Formally speaking, an ADS performance metric is a mapping
\begin{equation}\label{eq:metric}
    \mathcal{M}: \mathcal{D}_{O, k} \rightarrow G.
\end{equation}
The metric takes an extracted data set as input containing all necessary states characterizing the interactive motion between a certain SV, other road users, and driving environment, evolved with time. The metric then maps the input data set to a target set $G$ with $g \in G$ denoting a certain outcome that characterizes a certain performance-related property of the SV. 

Note the set $G$ comes with various physical properties for interpreting the performance of the SV $k$ interacting with other POVs in the ODD $O$ revealed by the extracted data set $\mathcal{D}_{O, k}$. The 33 base metrics (one base metric might have multiple variants with different selections of hyper-parameters as addressed later in Section~\ref{sec:performance_da}) studied in this paper admit metrics with properties including time (e.g., TTC~\cite{hayward1972near}), distance (e.g., DSS~\cite{okamura2011impact}), risk scalar (e.g., FMRI~\cite{kalra2016driving}), unit-less index (e.g., PSD~\cite{astarita2019traffic}), Boolean rating (e.g., RSS~\cite{shalev2017formal}), and various combinations of the above.

Moreover, the set $G$ also admits different mathematical forms. For example, TTC has a time value for each state $\s \in \mathcal{D}_{O,k}$, leading to the final output as a vector $g \in G = \R_{\geq0}^{N_s}$. Note that $N_s = |\mathcal{D}_{O,k}|$ denotes the number of states in (i.e., the cardinality of) $\mathcal{D}_{O,k}$. On the other hand, TA has a time value for each incident $\mathcal{I} \subset \mathcal{D}_{O,k}$, leading to a different output vector from the TTC case as $g\in G=\R_{\geq0}^{N_I}$, where $N_I$ denotes the total number of incidents in the extracted data set. Overall, majority of the metrics studied take the mathematical output of one of the following three forms: (i) a Real vector of dimension $N_s$, (ii) a Real vector of dimension $N_I$, or (iii) a Boolean vector of dimension $N_s$. Metrics of different output forms are not directly empirically comparable. With the same mathematical form as the output, metrics may still take different physical interpretations as mentioned earlier in Section~\ref{sec:intro}. This challenge can be overcome as discussed in Section~\ref{sec:performance_da} for direct performance comparisons.

Overall, the metrics considered in this paper are all \emph{leading measures} except for AD, AM, FMRI, CR, $\bar{\epsilon}\alpha$-ASS, Jerk, GT, and THW, which are \emph{lagging measures}, as defined in ~\cite{fraade2018measuring, griffor2019workshop}. If classified by the output unit, as in~\cite{elli2021evaluation,jammula2022evaluation}, among the 33 base metrics, eight metrics are time-based, and two metrics are distance-based in their original proposal. While the aforementioned features capture some of the differences among metrics, it is difficult to justify the acceptability of a certain metric by the way they are classified and compared in the literature. This inspires the constructive diversity analysis in the following section. 

\section{The Constructive Diversity Analysis}\label{sec:constructive_da}

This section introduces the proposed framework that identifies the constructive differences among the 33 base metrics by classifying them into three categories. For a detailed description of each metric mentioned in this section, one can refer to Appendix~\ref{apx:constructive_da:overview}. Note that terms reaction time, response time, and time delay, used within individual metrics are commonly referred to as response time in the remainder of this paper. Each of the 33 base metrics has various assumptions, working conditions, safety level justifications, and format of outputs. However, as described in the following discussion, these metrics can essentially be interpreted in a structurally consistent manner.

In particular, this paper proposes the following three types of algorithms that contribute to the studied metrics, including (i) the \emph{model-predictive} algorithm, (ii) the \emph{observation-transform} algorithm, and (iii) the \emph{statistical inference} algorithm. The three classes of algorithms are primarily based on (i) the form of assumptions, and (ii) the type of assurance provided through the metric outcome. 

\begin{table*}[!t]
    \caption{Some modeled and behavioral assumptions for SV and POVs, among all model-predictive metrics. The metric index admits the same configuration as defined in TABLE~\ref{tab:metrics}.}
    \label{tab:sv_assumption}
    \centering
    \resizebox{0.99\textwidth}{!}{%
    \begin{tabular}{l|l|c|c|c|c|c|c|c|c|c|c|c|c|c|c|c|c|c|c|c|c|c|c}
    \hline
    \ & Assumption & 1 & 2 & 3 & 4 & 5 & 6 & 7 & 8 & 9 & 10 & 11 & 12 & 13 & 14 & 15 & 16 & 17 & 18 & 19 & 20 & 21 & 22\\ \hline
    \multirow{5}{*}{SV} & Steady-state &\checkmark &\checkmark & \ & \ & \ & \ & \ & \ & \ &\checkmark & \ & \checkmark & \ & \checkmark &\checkmark & \ & \ & \checkmark &\ & \ & \ & \  \\ \cline{2-24}
    & Instantaneous-control &\ &\ &\checkmark&\ &\ &\ &\checkmark &\checkmark &\checkmark &\ &\checkmark &\ &\checkmark &\ &\ &\ &\ &\ &\checkmark & \ & \checkmark & \  \\ \cline{2-24}
    & Evasive-maneuver &\ &\ &\ & \checkmark &\checkmark &\checkmark &\ &\ &\ &\ &\ &\ &\ &\ &\ & \checkmark &\checkmark &\ &\ &\checkmark &\checkmark&\checkmark \\ \cline{2-24}
    & Response time considered &\ & \ & \ & \ & \checkmark & \checkmark & \ & \ & \ & \ & \ & \ & \ & \ &\ & \checkmark & \checkmark & \ &\ & \checkmark & \checkmark & \checkmark  \\ \cline{2-24}
    & Point-mass & \checkmark & \checkmark & \checkmark & \ & \checkmark & \checkmark & \checkmark & \checkmark & \checkmark & \checkmark & \checkmark & \checkmark & \checkmark & \checkmark &\checkmark & \checkmark & \checkmark & \checkmark & \checkmark & \checkmark & \checkmark &\checkmark  \\ \hline
     \multirow{3}{*}{POV} & Steady-state &\checkmark & \ & \ & \ & \ & \ & \checkmark & \checkmark & \ & \checkmark& \ & \checkmark & \checkmark & \checkmark & \checkmark & \ & \ & \checkmark & \ & \ & \ & \checkmark  \\ \cline{2-24}
    & Instantaneous-control &\ &\checkmark & \checkmark &\ &\ & \ & \ & \ & \checkmark & \ & \checkmark &\ &\ &\ &\ &\ &\ &\ &\checkmark & \ & \ & \  \\ \cline{2-24}
    & Worst-case &\ &\ &\ &\checkmark & \checkmark & \checkmark &\ &\ &\ &\ &\ &\ &\ &\ &\ &\checkmark & \checkmark& \ &\ & \checkmark & \checkmark & \ \\ \hline
    \end{tabular}%
    }
    \vspace{-5mm}
\end{table*}

\subsection{The Class of Model-Predictive Metrics}

The class of model-predictive metrics adapts the name from~\cite{weng2021class}, but the interpretation has been extended to a broader range of metrics. It determines the performance justification outcome based on a series of assumptions posed upon the dynamic and behavioral models of the SV and the POV. In this paper, 22 out of the 33 base metrics can be interpreted by the model-predictive perspective with the algorithm template shown in Algorithm~\ref{alg:mp}.

Consider some metrics studied in this paper as examples. (i) Time to Collision (TTC) creates a predictive motion trajectory assuming both vehicles are maintaining steady-state (i.e., maintaining the instantaneous velocity and heading angle indefinitely) and are following the linear double-integrator dynamics (also known as the point-mass model). The TTC is thus the time such a predictive trajectory takes to reach one of the collision states. The result is also provably valid (i.e., the SV is guaranteed to encounter a collision at the given TTC) under the given assumptions. One can refer to Fig. 1 in~\cite{weng2021class} for an illustrative example of this interpretation. Other metrics among the 22 studied model predictive metrics which share similar assumptions are Reciprocal Time to Collision (RTTC), Time to Accident (TA), Time Exposed TTC (TET), Time Integrated TTC (TIT), and Time to Collision Violation (TTCV). (ii) Rear End Collision Risk Index (RCRI) shares the same point-mass assumption with TTC. But the behavioral assumption is different as RCRI assumes both vehicles perform brake-to-stop (with an added response time assigned to the SV). The output of RCRI is also different from TTC as the metric gives a Boolean return based on whether a predictive collision would occur under the assumed models. Other metrics sharing akin assumptions are Potential Index for Collision with Urgent Deceleration (PICUD), Difference of Space Distance and Stopping Distance (DSS), and Time Exposed Rear End Collision Risk Index (TERCRI). (iii) Responsibility Sensitive Safety (RSS) expands the one-dimensional point-mass assumption used by TTC and RCRI to the two-dimensional plane with different behavioral assumptions assigned to the longitudinal and lateral interactions. In addition, Modified Time to Collision (MTTC), Required Longitudinal Acceleration (RLA), Brake Threat Number (BTN), and Modified Time to Collision Violation (MTTCV) share similar assumptions within them, while Deceleration Rate to Avoid the Crash (DRAC), Deceleration to Safety Time (DST), and Crash Potential Index (CPI) share same assumptions. Finally, not every metric takes the point-mass dynamics assumption. Model Predictive Instantaneous Safety Metric (MPrISM) and Criticality Metric~\cite{junietz2018criticality} (not studied within the 33 base metrics) consider the locally linearized bicycle kinematics model. Note that a seemingly minor difference among metrics in their constructive nature does not necessarily indicate that their empirical performance outcomes are close. One should observe this later in Fig.~\ref{fig:highd_ol_xx_times_t} where the minor difference regarding the predictive motion of the SV between TTC and MTTC leads to significant disagreement with a 0.67 Agreement Index.
\begin{algorithm}[H]
    \begin{algorithmic}[1]
    \State {\bf Assume:} motion models $f_0$ and $f_1$, behavioral models $b_0$ and $b_1$ for the SV and the POV
    \State {\bf Input:} $\mathcal{D}_{O, k}$
    \State {\bf Initialize:} {$R=\emptyset$, where R is a set of metric outcomes for each state/incident in $\mathcal{D}_{O, k}$}
    \State {\bf For} $\s$ in $\mathcal{D}_{O, k}$ {\bf do}
    \State {\ \ \ \ Start from $\s$, assume SV does $b_0$ and evolves with $f_0$, POV does $b_1$ and evolves with $f_1$}
    \State {\ \ \ \ $R$.\texttt{append}(a certain outcome of the above assumption)}
    \State {{\bf End For}}
    \State {{\bf Output:} $R$ or a certain transformation of $R$}
    \end{algorithmic}
    \caption{The model-predictive metric algorithm template } \label{alg:mp}
\end{algorithm}

In general, model-predictive metrics interpret the predictive safety outcome in various ways. Among the 22 model-predictive metrics studied in this paper, TTC, MTTC, Potential Time to Collision (PTTC), MPrISM, TERCRI, TET, TIT, and TA give a time-related output (through either a time-to-collision value or a certain unsafe time duration). PICUD and DSS determine the performance through the notion of distance to a certain unsafe outcome. RLA, DRAC, and DST derive the acceleration required to avoid a certain predictive collision. TTCV, MTTCV, Fuzzy Safety Model (FSM), RSS, UN Regulation 157 (UNReg157), and RCRI give Boolean outcomes. BTN, CPI, and RTTC give other types of outputs.

Moreover, seven metrics assume the SV maintains steady-state. Seven metrics assume a certain response time of the SV, yet the assumed vehicle behaviors within the response time are of many variants. Eight metrics assume the SV maintains instantaneous control action (immediately or after the response time if applicable). Twenty-one metrics are based on the point-mass dynamics (i.e., the double-integrator linear dynamics) assumption. Eight metrics account for the evasive maneuver (i.e., the evasive action for collision avoidance) for the SV. Seven metrics admit the worst-case (i.e., the optimal strategy to induce a potential collision) for the POV. A detailed list of commonly used behavioral and dynamics assumptions can be found in TABLE~\ref{tab:sv_assumption}.

Another remark regarding the class of model-predictive metrics is related to the assurance one provides through the outcomes. Thanks to the well-structured dynamics assumptions (mostly of linear form) and behavioral models (mostly time-invariant), many model-predictive metrics come with a certain level of so-called safety assurance. For example, RSS ``guarantees" the SV's absence from any collision for which the SV would take responsibility. MPrISM ``guarantees" that the SV will not collide with the POV within the predictive time horizon with certain inputs. It is important to recognize that such assurances and guarantees are typically only valid within the model predictive regime set by the assumptions mentioned above.

\subsection{The Class of Observation-Transform Metrics}
In contrast to the modeled nature of the aforementioned metrics, some metrics are \emph{assumption-free}, such as Crash Index (CI), Proportion of Stopping Distance (PSD), Aggressive Driving (AD), Accident Metric (AM), Jerk, Gap Time (GT), Time Headway (THW), Level of Unsafety (LU), Collision Rate (CR), as well as metrics not studied in detail within this paper, such as the disengagement rate~\cite{nowakowski2015development}. All of the above-mentioned metrics are direct transformations of a certain observation from the input data set. The correctness of the outcome is guaranteed without any posed assumptions.

Among the various observation-transform type metrics, there are constructively simple ones related to the frequency count of a certain event of interest such as CR (collision rate), AM, AD, and the disengagement rate. There are also metrics like LU where the metric outcome is a multiplication of several features intuitively related to the safety performance of vehicles.

Note that for both the model-predictive metrics and the observation-transform metrics, it is typically expected that the cardinality of the input data set, $N_s$ (or $N_I$), is sufficiently large (i.e., the more data the better). However, fundamentally speaking, the data size is mostly irrelevant to the assurance or accuracy provided through the metric output. This makes the above two classes of metrics different from the class of statistical-inference metrics described later.

\subsection{The Statistical-Inference Metrics} 
In many ways, the statistical-inference metrics, including $\bar{\epsilon}\alpha$-Almost Safe Set ($\bar{\epsilon}\alpha$-ASS) and Failure-free Miles Risk Inference (FMRI) studied in this paper, and the observation-transform metrics are similar as statistical inference is, to a certain extent, a special kind of transform.

On the other hand, the statistical-inference metrics are also uniquely different as all such metrics are \emph{probabilistic complete}. That is, (i) the data size matters and affects the outcome, and (ii) as $N_s$ (or $N_I$) tends to infinity, the probability of obtaining an assured safety outcome always tends to one. Moreover, the assured safety outcome is an unbiased generalization of the SV's safety performance from the observed data set to the unobserved cases within the same ODD.

In practice, FMRI requires consecutive observation of safe driving mileage over an implicitly defined ODD revealed through the input data set. $\bar{\epsilon}\alpha$-ASS does not have the restriction of safe driving data only, and it is also domain-specific as it characterizes the specific ODD within which the SV is safe with a certain probability, revealed through the data. 

\section{The Performance Diversity Analysis}\label{sec:performance_da}
In general, the empirical diversity analysis in this section shows that \emph{the metrics heavily disagree with each other}. As a result, the common practice of demonstrating the utility of a safety metric proposal through empirical comparisons may be problematic.

The demonstration starts with the introduction of the two raw data sets used. The agreement justification is introduced next as the main tool to perform the performance diversity analysis among most of the metrics. A description of metric variants, due to different values of hyper-parameters within a metric, is then provided. This is followed by a case study exhibiting the application of metrics to a sample extracted data set, and insights on observed results, capturing the earlier described constructive diversities. Finally, various empirical outcomes are presented.

\subsection{Data set construction and processing}

\begin{figure*}[!ht]
\centering
\begin{subfigure}{0.32\linewidth}
  \centering
  \includegraphics[trim={2cm 1cm 1cm 1cm},clip,width=\textwidth]{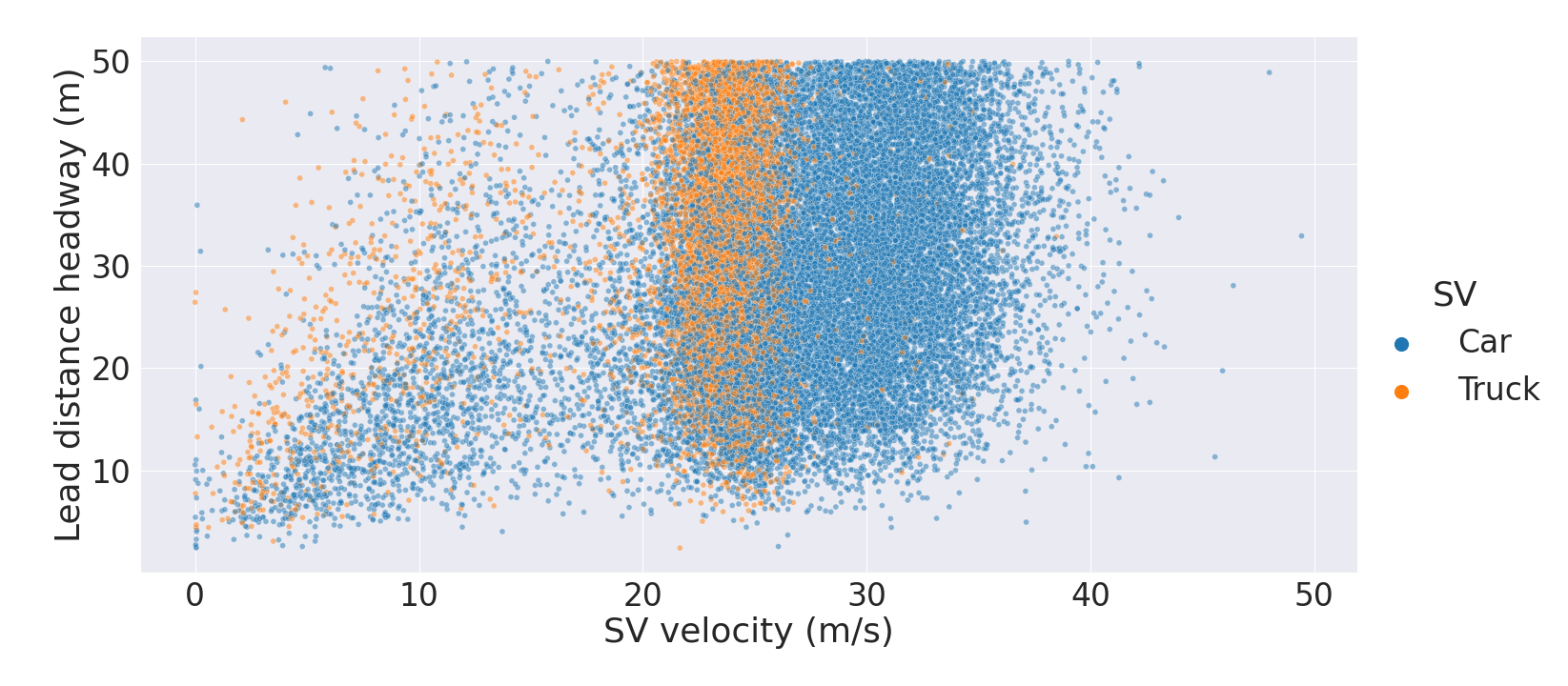}
  \caption{A uniformly sampled subset (10\%) of states from $\mathcal{D}^H$ with the leading-vehicle distance headway shorter than 50-metre.}
  \label{fig:highd_state}
\end{subfigure}%
\hspace{1em}%
\begin{subfigure}{0.32\linewidth}
  \centering
  \includegraphics[trim={2cm 1cm 1cm 1cm},clip,width=\textwidth]{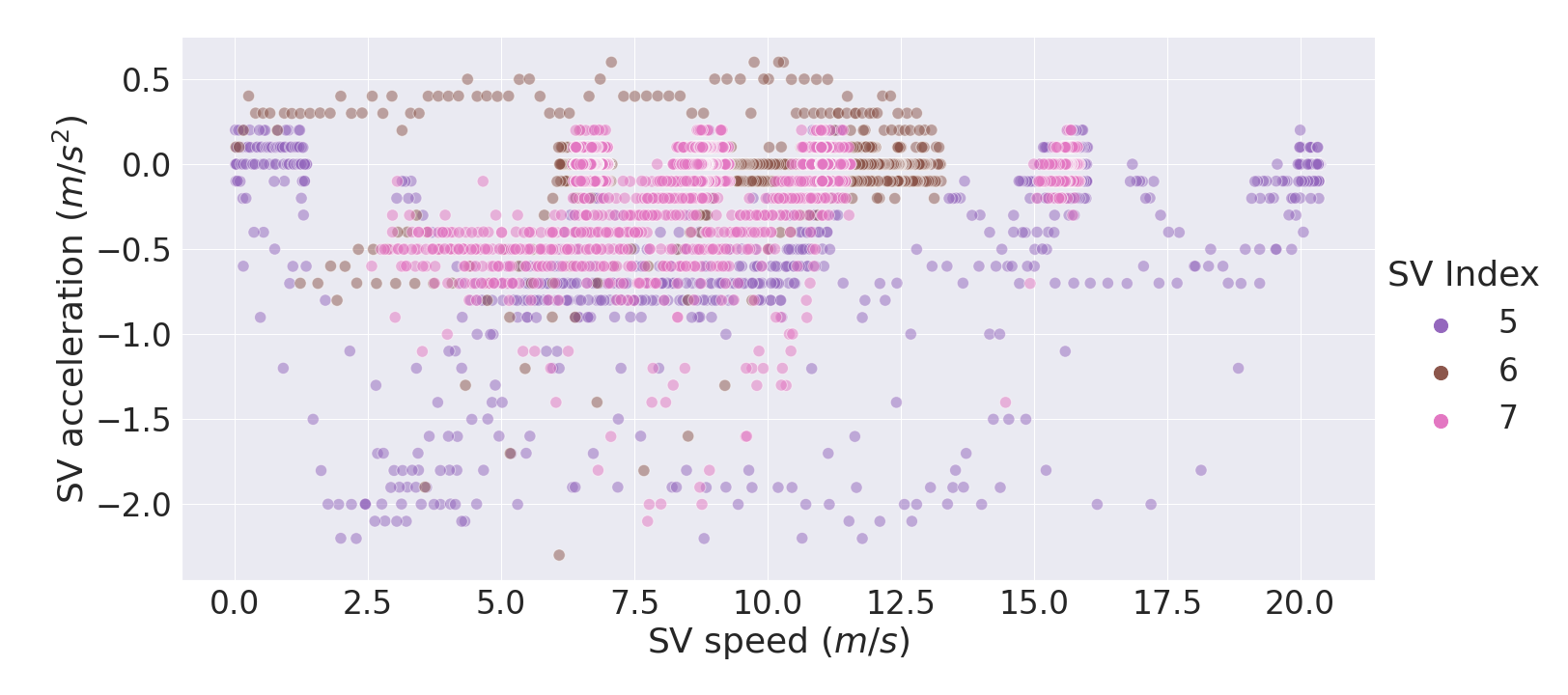}
  \caption{An overview of all to-collision states in $\mathcal{D}^V$ (i.e., all states in $\mathcal{D}^V$ that are about to encounter a vehicle-to-vehicle impact).}
  \label{fig:vice_state}
\end{subfigure}%
\hspace{1em}%
\begin{subfigure}{0.32\linewidth}
  \centering
  \includegraphics[trim={2cm 1cm 1cm 1cm},clip,width=\textwidth]{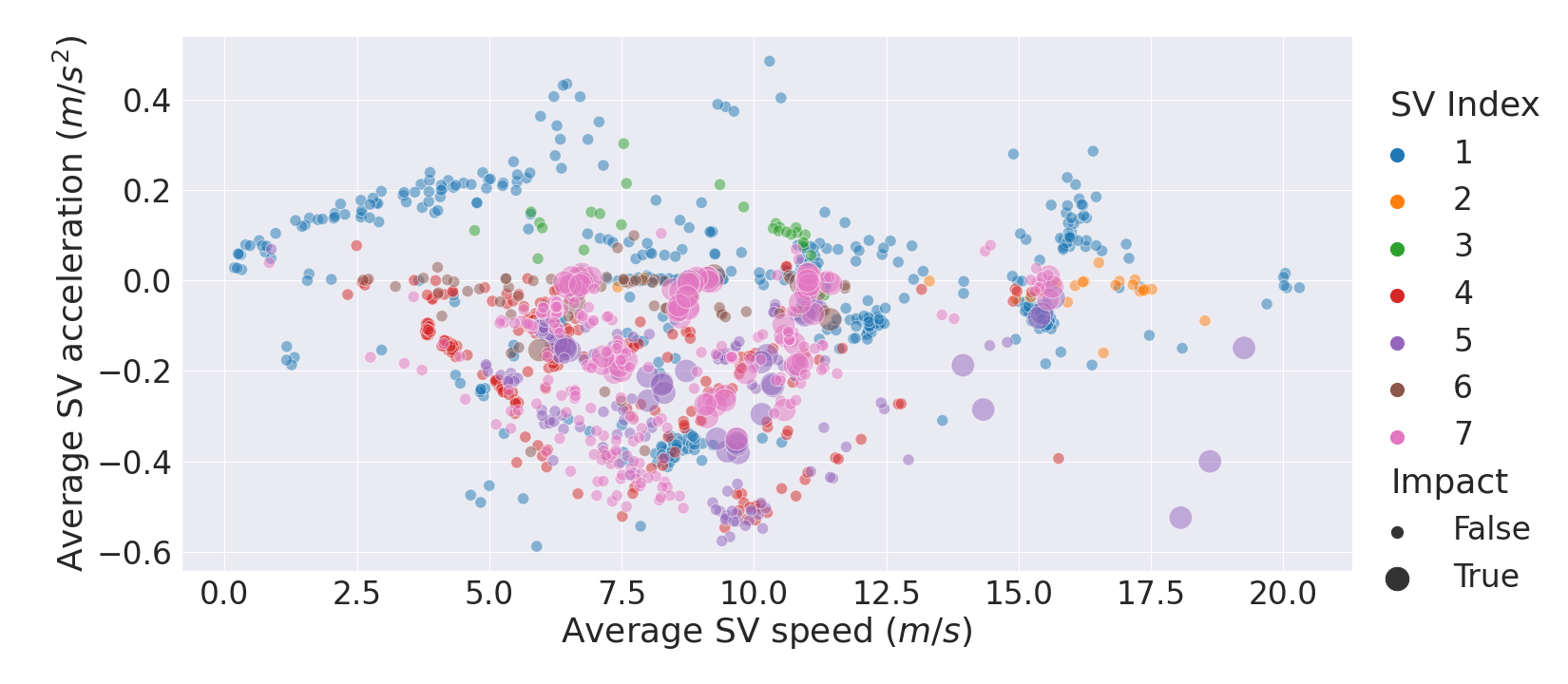}
  \caption{An overview of some of the statistical properties of all incidents from the VICE data set (i.e., all $I \subset \mathcal{D}^V$).}
  \label{fig:vice_incident}
\end{subfigure}
\caption{Some extracted properties of the HighD and the VICE data sets. Moreover, within the 1366 incidents in the VICE data set, 22\% of the incidents have minimum DHW less than two-metre, including 117 incidents with collision events.}
\label{fig:dataset}
\vspace{-3mm}
\end{figure*}
% 190 less than 1, 300 less than 2, 411 less than 3, 553 less than 4, 682 less than 5

\subsubsection{The HighD Data Set}\label{sec:data:highd}
The HighD data set~\cite{krajewski2018highd} (including the HighD and the HighD Plus) is a data set of naturalistic vehicle trajectories recorded on German highways. The data set includes a mixture of cars and trucks operating on straight road highway segments. The particular HighD data set, $\mathcal{D}^H$, used in this study is a subset of the raw HighD and the HighD Plus, where all vehicles in each track are iteratively treated as the SV, provided they are in the rightmost or second from the rightmost lane in both driving directions. For each SV, the corresponding data set is extracted at a frequency of 5-Hz, which includes global and local positions, velocities, accelerations, heading angles, lengths, widths, etc. The obtained result in $O_l$, comprising of the follower SV and the leading POV, is thus denoted as $\mathcal{D}^H_{O_l}$.

\subsubsection{The VICE Data Set}\label{sec:data:vice}
The Vehicle Impending Crash Event (VICE) data set was developed at the National Highway Traffic Safety Administration (NHTSA)'s Vehicle Research and Test Center. It includes 1554 high-frequency (50 Hz), high-accuracy, multi-vehicle (2 to 4 agents) events with various crash-imminent interactions. They were extracted from closed-course vehicle tests performed at the Transportation Research Center between 2017 and 2019. The data set involves 7 anonymous SVs, denoted by an integer index from 1 to 7. Six of them are ADAS-equipped vehicles commercially available in U.S. market. The corresponding extracted data set in $O_l$, which includes global and local positions, velocities, accelerations, heading angles, lengths, widths, etc., similar to the HighD extracted data set, is thus denoted as $\mathcal{D}^V_{O_l, i}$ for car $i$ in VICE. Moreover, let $\mathcal{D}^V_{O_l} = \bigcup_{i\in\Z_7}\mathcal{D}^V_{O_l, i}$.

While the HighD data set is a representative example of naturalistic driving data with mostly low risk behaviors (e.g., the data set involves zero collision events), the VICE data set involves many high-risk vehicle-to-vehicle interactions, as the closed-course vehicle tests that contribute to VICE all follow NHTSA draft research test procedures for a variety of ADAS technologies. Such a fundamental behavioral difference of vehicles contributes significantly to the diversity analysis of a broad spectrum of ADS performance metrics as we should present later.

One can refer to Fig.~\ref{fig:dataset} for some extracted statistical properties of the two described data sets.

\subsection{The Agreement Justification}\label{sec:agreementjustification}

As mentioned earlier, the key challenge of performing empirical diversity analysis w.r.t. various metrics lies in the heterogeneity of the output space $G$. That is, the metrics' outputs are of different ranges and different units with different physical interpretations. W.l.o.g., consider two metrics $\mathcal{M}_1: \mathcal{D}_{O, k} \rightarrow G_1$ and $\mathcal{M}_2: \mathcal{D}_{O, k} \rightarrow G_2$. The diversity analysis between $\mathcal{M}_1$ and $\mathcal{M}_2$ generally follows a two-step procedure, including (i) transferring the safety metric to a classifier, and (ii) applying various agreement justification methods w.r.t. the classification outcomes.

For metrics admitting Boolean value outputs (i.e., metric outputs of the form $\mathbb{B} \times N_s$ or $\mathbb{B} \times N_I$), they are inherently classifiers by their functional nature as each individual state (or incident) is classified into one of the two Boolean outputs. Moreover, consider metrics that admit the subsets of $\R \times N_s$ (or $\R \times N_I$) as the outputs. Note that all such metrics studied in this paper admit the unique property that the metric output is either monotonic or consists of elements that are monotonic w.r.t. a certain notion of safety performance justification (e.g., a large/low TTC value implies low/high risk). W.l.o.g., for such metrics, let the smaller metric output imply higher risk. Consider ${\mathcal{D}_{O, k} \choose 2}$ as the set of all pairwise combinations of all points in $\mathcal{D}_{O, k}$. For any $(\s_i, \s_j) \in {\mathcal{D}_{O, k} \choose 2}$ and $m \in \{1,2\}$, let
\begin{equation}\label{eq:real_to_class}
    \bar{\mathcal{A}}_{ij}^{m}:=\bar{\mathcal{A}}(\mathcal{M}_m, \s_i, \s_j) = \text{sign}(\mathcal{M}_m(\s_i)-\mathcal{M}_m(\s_j)).
\end{equation}
The above equation transfers the metric outputs of Real values to a three-class classification outcome in $\{-1, 0, 1\}$. As a result, $\mathcal{M}_1$ agrees with $\mathcal{M}_2$ regarding the safety performance of $\s_i$ and $\s_j$ if they both consider $\s_i$ being safer than ($1$) or equally safe with ($0$) or of higher risk than ($-1$) $\s_j$. A similar notion directly generalizes to the incident-based analysis as well, hence is omitted.

To this end, the pairwise agreement among most of the metrics considered in this paper can be studied as a classifier agreement evaluation problem which admits various evaluation methods like Precision-Recall~\cite{he2009learning}, Area under the receiver operating characteristic (ROC) curve (AUC)~\cite{bradley1997use}, Mathews Correlation coefficient (MCC)~\cite{matthews1975comparison}, Cohen's Kappa coefficient~\cite{mchugh2012interrater}, Index of Agreement~\cite{willmott1981validation}, to name a few. Different methods capture different perspectives of the outcome agreement, each with its own set of advantages and disadvantages. In particular, we consider Agreement Index (AID) and Precision-Recall scores in this study as they are suitable for the problem presented in this paper with limited number of classes. W.l.o.g., let $\mathcal{C}^m_{O, k}$ ($m\in\{1,2\}$) be the set of classification outcomes transferred from using metric $\mathcal{M}_m$ analyzing the extracted data set of $\mathcal{D}_{O, k}$.

AID $\mathcal{A}$ between comparable metrics $\mathcal{M}_1$ and $\mathcal{M}_2$ is henceforth defined as
\begin{equation}\label{eq:aid}
    \mathcal{A}(\mathcal{M}_1, \mathcal{M}_2, \mathcal{D}_{O, k})=\frac{\sum_{i=1}^{|\mathcal{C}^1_{O, k}|}\llbracket \mathcal{C}^1_{O, k}[i],\mathcal{C}^2_{O, k}[i] \rrbracket}{|\mathcal{C}^1_{O, k}|}.
\end{equation}

For the precision and recall study in case of Boolean output metrics, the \texttt{True} classified outputs are considered as positives (i.e., the class of interest). The methodology followed in general can be applied to the \texttt{False} classified outputs as well in a similar manner. Considering $\mathcal{C}^1_{O, k}$, a set of transformed classification results of a Boolean output metric $\mathcal{M}_1$, to be compared with $\mathcal{C}^2_{O, k}$, another set of transformed classifier results of another Boolean output metric $\mathcal{M}_2$. A classified output in $\mathcal{C}^1_{O, k}$ is considered correct if it matches with the corresponding actual output in $\mathcal{C}^2_{O, k}$. True Positives (TPs), False Positives (FPs), and False Negatives (FNs) are thus defined as the number of correct classified positive outputs in $\mathcal{C}^1_{O, k}$, incorrectly classified positive outputs in $\mathcal{C}^1_{O, k}$, and incorrectly classified negative outputs in $\mathcal{C}^1_{O, k}$, respectively. Henceforth, the precision measures the proportion of total predicted positives that are relevant~\cite{buckland1994relationship}. The recall measures the proportion of total positives which are correctly classified~\cite{buckland1994relationship}. The measures defined earlier (i.e., TPs, FPs, and FNs) are used to calculate these scores~\cite{powers2011evaluation}. The precision score is a ratio of TPs and the sum of TPs and FPs, while the recall score is a ratio of TPs and the sum of TPs and FNs. These scores vary from 0 to 1. A high value for both implies a classifier returning accurate results (high precision score), as well as returning a majority of positive results (high recall score)~\cite{buckland1994relationship}. Specifically for precision results shown in this paper, a score of 1 indicates all classified positive outputs in $\mathcal{C}^1_{O, k}$ agree with their corresponding outputs in $\mathcal{C}^2_{O, k}$, and a score of 0 indicates none of the classified positive outputs in $\mathcal{C}^1_{O, k}$ agree with their corresponding outputs in $\mathcal{C}^2_{O, k}$. It is noted here that given the methodology introduced, precision acts as a special case of AID calculation that was introduced earlier.

For example, the precision score defined w.r.t. the class $c$ of positive outputs (for Boolean output metrics, $c$ can be either \texttt{True} or \texttt{False}) comparing $\mathcal{M}_1$ against $\mathcal{M}_2$ is defined as
\begin{equation}\label{eq:precision_c}
    \mathcal{P}_c(\mathcal{M}_1, \mathcal{M}_2, \mathcal{D}_{O, k})\!=\!\frac{\sum_{i=1}^{|\mathcal{C}^1_{O, k}|}\llbracket \mathcal{C}^1_{O, k}[i],\mathcal{C}^2_{O, k}[i] \rrbracket\!\cdot\!\llbracket \mathcal{C}^1_{O, k}[i],c \rrbracket}{\sum_{i=1}^{|\mathcal{C}^1_{O, k}|}\llbracket \mathcal{C}^1_{O, k}[i],c \rrbracket}.
\end{equation}

In the above equation, the numerator represents the TPs, while the denominator represents the sum of TPs and FPs, when comparing $\mathcal{C}^1_{O, k}$ against $\mathcal{C}^2_{O, k}$. Note $\mathcal{P}_c(\mathcal{M}_1, \mathcal{M}_2, \mathcal{D}_{O, k})$ is not necessarily equivalent with $\mathcal{P}_c(\mathcal{M}_2, \mathcal{M}_1, \mathcal{D}_{O, k})$, as we shall also observe empirically in Section~\ref{sec:performance_da:empirical}. For the case of Real value output metrics (i.e., metrics that admit the subsets of $\R \times N_s$ (or $\R \times N_I$) as the outputs) transformed to a three-class classification outcome, ``micro" averaging is used for precision score calculation~\cite{grandini2020metrics}. It is highlighted here that micro average precision score results are same as AID values, as established later through Fig.~\ref{fig:vice_ol_xx_times_i_mp_hm} and Fig.~\ref{fig:vice_ol_xx_times_i}, respectively.

\subsection{Metric Variants} \label{sec:metric_variants}

The hyper-parameter values used within a metric result in different metric variants that are used for the empirical study. For the remainder of this paper, these variants are typically determined by a three-step procedure. (i) For the hyper-parameter values already present in the metric proposal references, we have respected the defined values and used them as they are. (ii) Otherwise, we have stated the reference for the used values which are mostly based on standard testing procedures~\cite{euro2017test}. (iii) For the hyper-parameters in data set based metric variants, we have determined the hyper-parameters from the analysis of the HighD and the VICE data sets, respectively. Metric variants inspired by metric references and test procedures are distinguished by a numeric character (e.g., PICUD1 and PICUD2). Metric variants inspired by data sets admit the naming routine as V (for hyper-parameter values derived from the VICE data set) and H (for hyper-parameter values derived from the HighD data set) appended to the acronym of the metric (e.g., PICUD\_H and PICUD\_V).

From the analysis performed on the HighD data set, a maximum deceleration value of 4.4~$\mathrm{m/s^2}$ for cars and 3.2~$\mathrm{m/s^2}$ for trucks is used as hyper-parameter in the HighD specific metric variants (i.e., PICUD\_H, BTN\_H, CPI\_H, RCRI\_H, and TERCRI\_H). While for the VICE data set based metric variants, a maximum deceleration value of 4~$\mathrm{m/s^2}$ is used (i.e., PICUD\_V, BTN\_V, CPI\_V, RCRI\_V, and TERCRI\_V).

Apart from the data set inspired choices of hyper-parameters, other metric variants are also parameterized following the three-step procedure mentioned above. PICUD1 and PICUD2 involve a maximum deceleration value of 3.3~$\mathrm{m/s^2}$~\cite{uno2002microscopic} and 6~$\mathrm{m/s^2}$~\cite{seiniger2015methodology}, respectively, and a reaction time of one-second in both. DSS assumes a reaction time of 1.08-second~\cite{okamura2011impact}, while the deceleration value is determined by the multiplication between the coefficient of friction (0.7 from ~\cite{navin1996vehicle}) and the acceleration due to gravity (9.81~$\mathrm{m/s^2}$). DST requires a safety time gap value of 1.4-second~\cite{okamura2011impact} to be maintained w.r.t the leading POV. BTN1 and BTN2 use maximum deceleration value of 9.82~$\mathrm{m/s^2}$~\cite{Jansson2005CollisionAT} and 6~$\mathrm{m/s^2}$~\cite{seiniger2015methodology}, respectively. CPI1 and CPI2 use maximum deceleration value of 8.45~$\mathrm{m/s^2}$~\cite{cunto2008assessing} and 6~$\mathrm{m/s^2}$~\cite{seiniger2015methodology}, respectively. RCRI1 and RCRI2 involve a maximum deceleration value of 3.4~$\mathrm{m/s^2}$~\cite{oh2006method} and 6~$\mathrm{m/s^2}$~\cite{seiniger2015methodology}, respectively, and a reaction time of 0.1-second~\cite{oh2006method} for both. TERCRI1 and TERCRI2 share the same maximum deceleration values as RCRI1 and RCRI2, respectively. A threshold TTC of three-second~\cite{hirst1997format} is used in TTCV, MTTCV, TET, and TIT. Also, MPrISM is used with two variants in Section~\ref{sec:performance_da}. The first variant gives an outcome of either (i) a predictive time ($\leq$ 1) within which the SV is expected to encounter a collision, or (ii) a statement that the SV is expected to be safe within the one-second predictive time-window. The second variant simplifies the first one with a Boolean outcome by making (i) as \texttt{False} and, (ii) as \texttt{True}. Finally, the hyper-parameter values for RSS variants are mentioned in TABLE~\ref{tab:rss_variants}. Note that RSS1 and RSS2 are parametrized based on calibration results from the naturalistic driving data w.r.t. Method-1 and Method-2 discussed by Huang et al.~\cite{huang2021rss}. Among the various metric variants within the class of observation-transform metrics, PSD\_V, PSD\_H, LU\_V, and LU\_H derive the maximum deceleration hyper-parameter value from the respective data sets. PSD uses a maximum deceleration of 6~$\mathrm{m/s^2}$~\cite{seiniger2015methodology}. LU1 and LU2 use a maximum deceleration value of 9.82~$\mathrm{m/s^2}$~\cite{Jansson2005CollisionAT} and 6~$\mathrm{m/s^2}$~\cite{seiniger2015methodology}, respectively.

\subsection{Metrics Application Demonstration}

This section demonstrates how some of the metrics can be applied to an extracted data set. The examples shown also capture metric formulation insights from Section~\ref{sec:constructive_da}.

\begin{figure}[!t]
    \vspace{2mm}
    \centering
    \includegraphics[trim={0cm 0cm 0.75cm 0cm},clip,width=.48\textwidth]{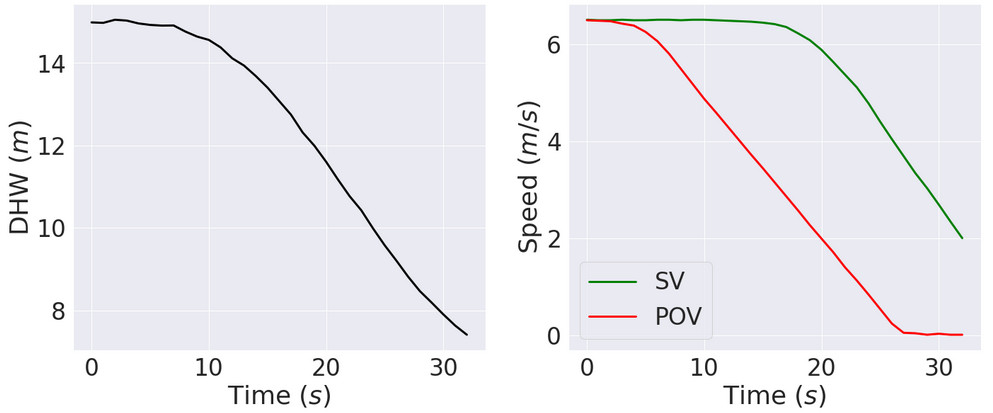}
    \caption{An incident example extracted from the VICE data set, where the follower SV approaches a leading POV.}
    \label{fig:case_study_dynamics}
\end{figure}
\begin{figure*}[!h]
\vspace{2mm}
\centering
\begin{subfigure}{0.23\linewidth}
  \centering
  \includegraphics[trim={0cm 0cm 0cm 0cm},clip,width=\textwidth]{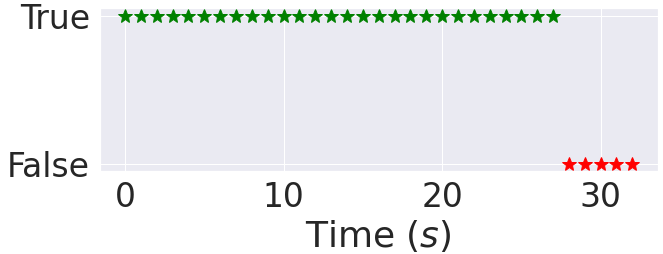}
  \caption{\footnotesize{MPrISM.}}
  \label{fig:case_study_metrics:mprism}
\end{subfigure}%
\hspace{1em}%
\begin{subfigure}{0.23\linewidth}
  \centering
  \includegraphics[trim={0cm 0cm 0cm 0cm},clip,width=\textwidth]{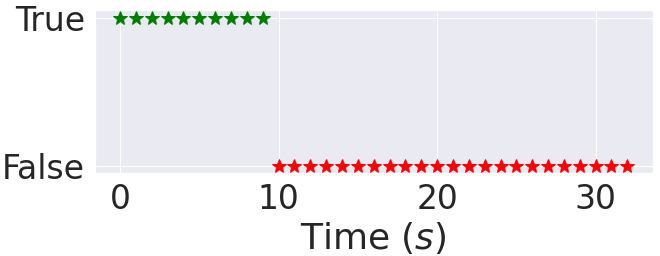}
  \caption{\footnotesize{FSM.}}
  \label{fig:case_study_metrics:fsm}
\end{subfigure}%
\hspace{1em}%
\begin{subfigure}{0.23\linewidth}
  \centering
  \includegraphics[trim={0cm 0cm 0cm 0cm},clip,width=\textwidth]{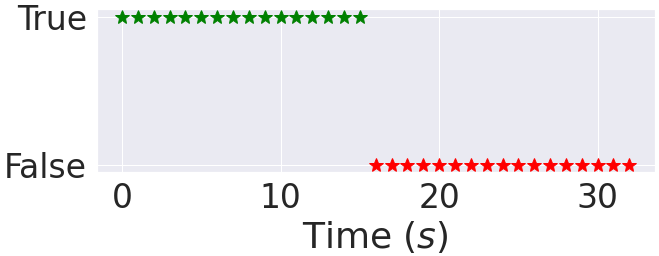}
  \caption{\footnotesize{RSS2.}}
  \label{fig:case_study_metrics:rss2}
\end{subfigure}
\hspace{1em}%
\begin{subfigure}{0.23\linewidth}
  \centering
  \includegraphics[trim={0cm 0cm 0cm 0cm},clip,width=\textwidth]{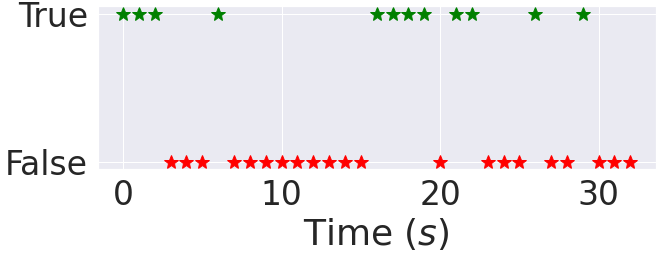}
  \caption{\footnotesize{MTTCV.}}
  \label{fig:case_study_metrics:mttcv}
\end{subfigure}%
\newline
\begin{subfigure}{0.32\linewidth}
  \centering
  \includegraphics[width=\textwidth]{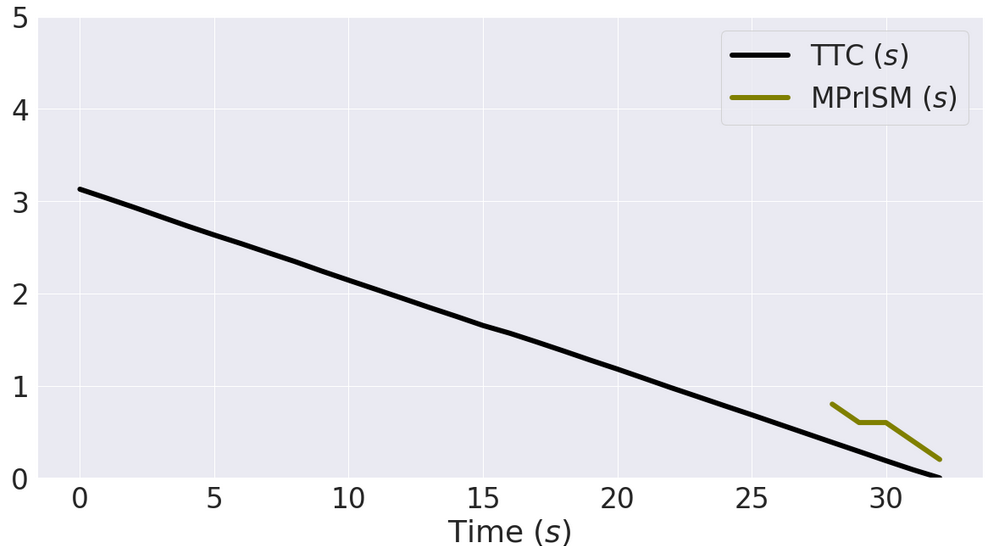}
  \caption{\footnotesize{ TTC and MPrISM.}}
  \label{fig:case_study_metrics:time}
\end{subfigure}%
\hspace{1em}%
\begin{subfigure}{0.32\linewidth}
  \centering
  \includegraphics[width=\textwidth]{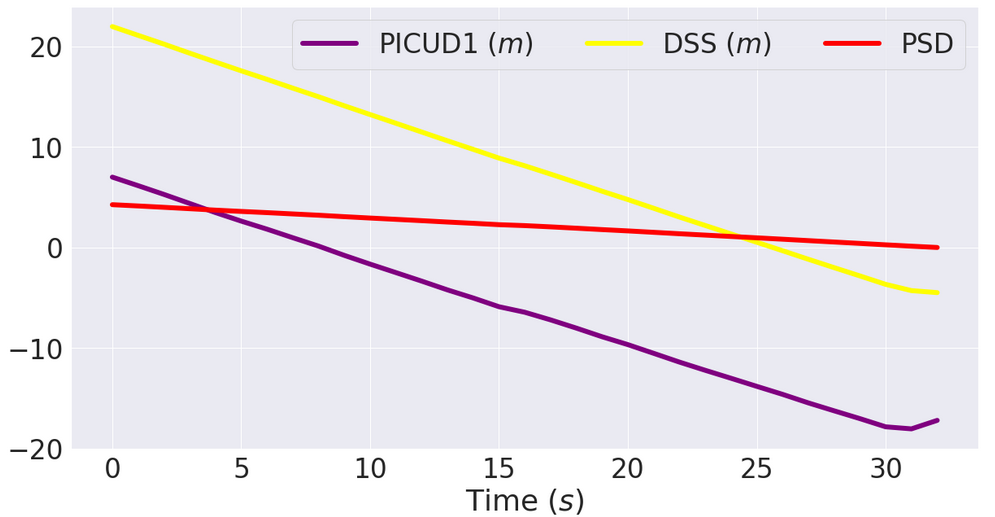}
  \caption{\footnotesize{PICUD, DSS, and PSD.}}
  \label{fig:case_study_metrics:dis}
\end{subfigure}
\begin{subfigure}{0.32\linewidth}
  \centering
  \includegraphics[width=\textwidth]{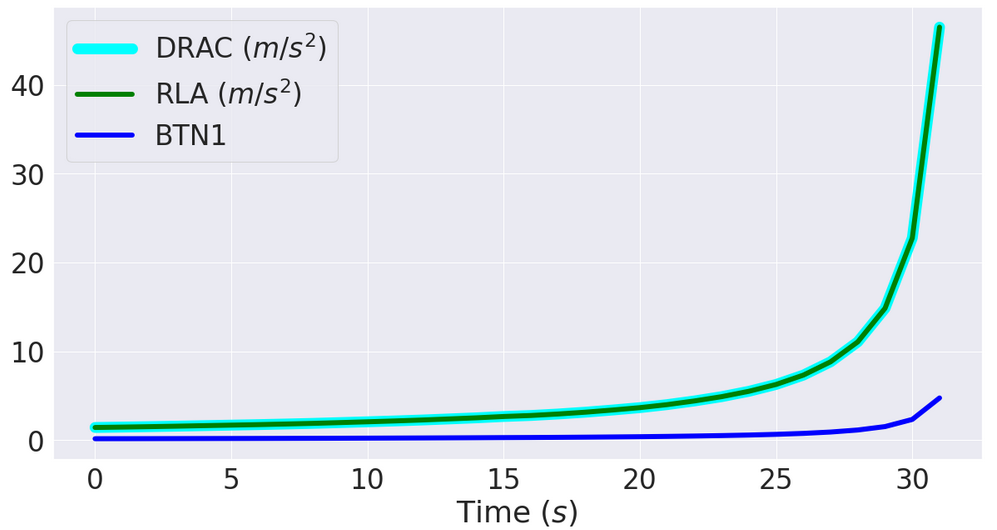}
  \caption{\footnotesize{DRAC, RLA, and BTN.}}
  \label{fig:case_study_metrics:acc}
\end{subfigure}%
\caption{Some metric outcomes w.r.t. the incident example shown in Fig.~\ref{fig:case_study_dynamics}.}
\label{fig:case_study_metrics}
\vspace{-3mm}
\end{figure*}
\begin{table}[]
    \caption{Hyper-parameters for RSS metric variants.}
    \label{tab:rss_variants}
    \centering
    \resizebox{0.49\textwidth}{!}{%
    \begin{tabular}{l|c|c|c}
    \hline
    Metric & RSS1 & RSS2 & RSS3\\ \hline
    Response Time (second) & 1.924 & 0.117 & 0.75  \\ \hline
    Maximum SV acceleration during response time ($\mathrm{m/s^2}$) & 3.805 & 4.836 & 3.805  \\ \hline
    Maximum leading POV deceleration ($\mathrm{m/s^2}$) & 4.585 & 8.086 & 7\\ \hline
    Maximum SV deceleration ($\mathrm{m/s^2}$) & 4.585 & 7.986 & 6 \\ \hline
    \end{tabular}%
    }
\end{table}

Fig.~\ref{fig:case_study_dynamics} illustrates the state transition of a sample incident $\mathcal{I}$ extracted from the data set $\mathcal{D}^V_{O_l, 7}$. Metric outcomes obtained from this incident are shown in Fig.~\ref{fig:case_study_metrics}. It is noted here that Fig.~\ref{fig:case_study_metrics:mprism} corresponds to the MPrISM variant with Boolean outcomes, and Fig.~\ref{fig:case_study_metrics:time} corresponds to the MPrISM variant with model predictive time to collision values, as established in Section~\ref{sec:metric_variants}.

Some of the key insights based on the constructive formulation established earlier are emphasized here. (i) From Fig.~\ref{fig:case_study_metrics:time}, MPrISM shows a low predictive time to collision compared to actual observation in the incident (no collision). This is due to much higher maximum deceleration limit of POV assumed in MPrISM (i.e., 6~$\mathrm{m/s^2}$) as compared to the actual maximum deceleration values of the involved leading POV (persistently less than 1~$\mathrm{m/s^2}$). This notable difference in the predictive behavior and the actual observed behavior serves as an example to highlight the discrepancy between the Model-Predictive construction and the Observation-Transform setup. (ii) Within the domain of model-predictive metrics, Fig.~\ref{fig:case_study_metrics:time} also highlights the differences between TTC and MPrISM, attributed to the different assumed behaviors of the follower SV and the leading POV under the two metrics. TTC assumes steady-state for both vehicles while MPrISM assumes worst-case POV maneuver and optimal collision avoidance strategy for the SV. (iii) From Fig.~\ref{fig:case_study_metrics:rss2} and Fig.~\ref{fig:case_study_metrics:mprism}, it is observed that RSS2 predicts unsafe outcomes comparatively earlier than MPrISM. Though the assumed behavior are close between RSS2 and MPrISM (i.e., both vehicles decelerate with maximum deceleration), RSS2 involves a response time of 0.117-second within which the SV accelerates and the POV decelerates with maximum deceleration. The assumed maximum deceleration limit of POV in RSS2 (8.086~$\mathrm{m/s^2}$) is also higher than the one assumed for MPrISM (6~$\mathrm{m/s^2}$). As a result, the higher deceleration rate coupled with distance covered during initial response time justifies the unsafe outcomes resulting earlier in RSS2. This underlines the differences among metric outcomes, contributed by the formulation differences. (iii) From Fig.~\ref{fig:case_study_metrics:fsm} and Fig.~\ref{fig:case_study_metrics:rss2}, FSM predicts unsafe outcomes prior to RSS2. This is justified based on the comfortable deceleration (i.e., 3~$\mathrm{m/s^2}$) used in the case of FSM at an earlier stage to avoid collision instead of maximum deceleration at a later time. Moreover, the maximum deceleration limit for the SV is assumed to be 6~$\mathrm{m/s^2}$, compared to 7.986~$\mathrm{m/s^2}$ in case of RSS2. FSM also assumes a longer response time (0.75-second) compared to RSS2. The differences in assumed behaviors and hyper-parameter values lead to unsafe outcomes earlier in FSM as compared to RSS2. (iv) Fig.~\ref{fig:case_study_metrics:mttcv} highlights the significant sensitivity of MTTCV to instantaneous acceleration values, resulting in fluctuations of safety outcomes. (v) Fig.~\ref{fig:case_study_metrics:dis} shows the results for PICUD1, DSS and PSD, all of which are based on DHW and stopping distances. Hence they all exhibit a trend similar to TTC. (vi) DRAC, RLA and BTN in Fig.~\ref{fig:case_study_metrics:acc} are acceleration based metrics following an inverse trend to TTC.

To summarize, the results for metrics in Fig.~\ref{fig:case_study_metrics:mprism}, Fig.~\ref{fig:case_study_metrics:fsm}, Fig.~\ref{fig:case_study_metrics:rss2}, Fig.~\ref{fig:case_study_metrics:mttcv} are strongly affiliated with the dynamics and behavioral assumptions of models involved. This results in significant disagreement observed among results, influenced heavily by the parameterized values of assumed maximum deceleration and response time. The metrics shown in Fig.~\ref{fig:case_study_metrics:time}, Fig.~\ref{fig:case_study_metrics:dis}, and Fig.\ref{fig:case_study_metrics:acc} follow a similar trend among each other. In this case, the predictive behavior does not play a determinant role in metric output. Hence the resulting dominant outcomes are mostly controlled by directly observable states related to distance and velocity.

\subsection{The Empirical Diversity Analysis}\label{sec:performance_da:empirical}

\begin{figure*}[!h]\label{fig:pairwiseaidcomparisons}
\vspace{2mm}
\centering
\begin{subfigure}{0.45\linewidth}
  \centering
  \includegraphics[trim={1cm 4cm 5cm 5cm},clip,width=\textwidth]{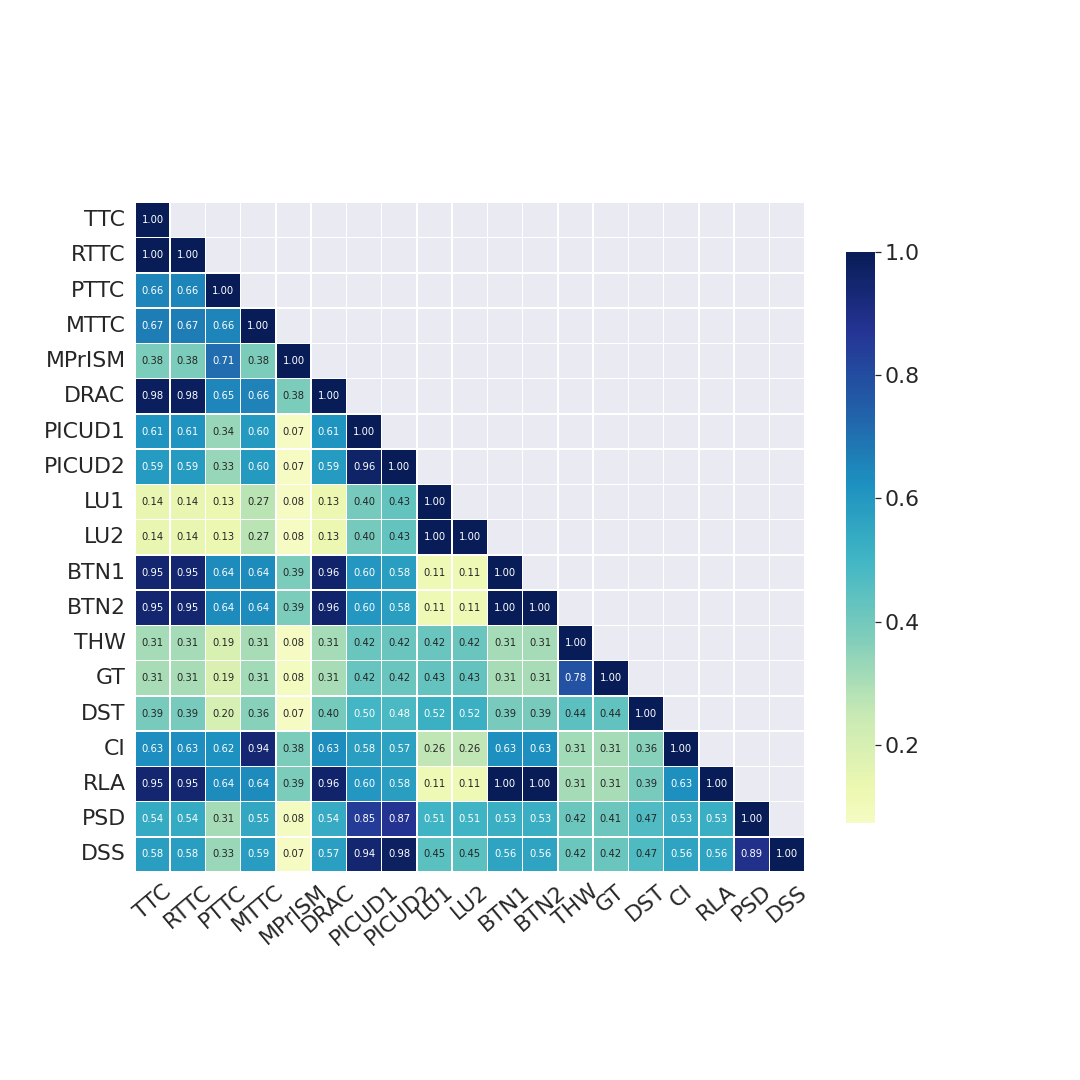}
  \caption{\footnotesize{Pairwise AIDs comparisons for metrics mapping $\mathcal{D}_{O_l}^H \rightarrow \mathbb{R}\times N_s$}.}
  \label{fig:highd_ol_xx_times_t}
\end{subfigure}
\hspace{1em}%
\begin{subfigure}{0.45\linewidth}
  \centering
  \includegraphics[trim={1cm 4cm 5cm 5cm},clip,width=\textwidth]{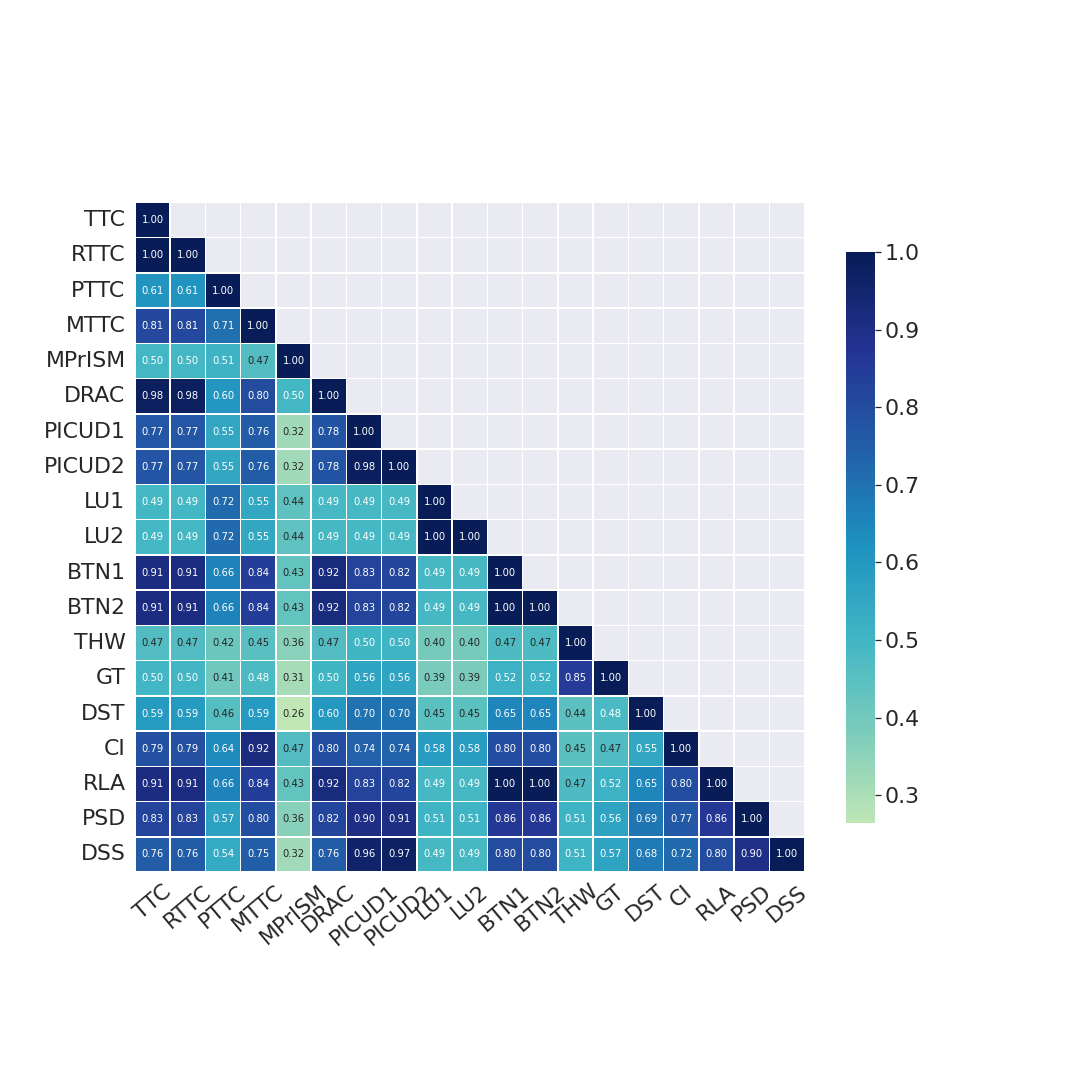}
  \caption{\footnotesize{Pairwise AIDs comparisons for metrics mapping $\mathcal{D}_{O_l}^V \rightarrow \mathbb{R}\times N_s$}.}
  \label{fig:vice_ol_xx_times_t}
\end{subfigure}%
\newline
\begin{subfigure}{0.30\linewidth}
  \centering
  \includegraphics[trim={0cm 4.5cm 5cm 5cm},clip,width=\textwidth]{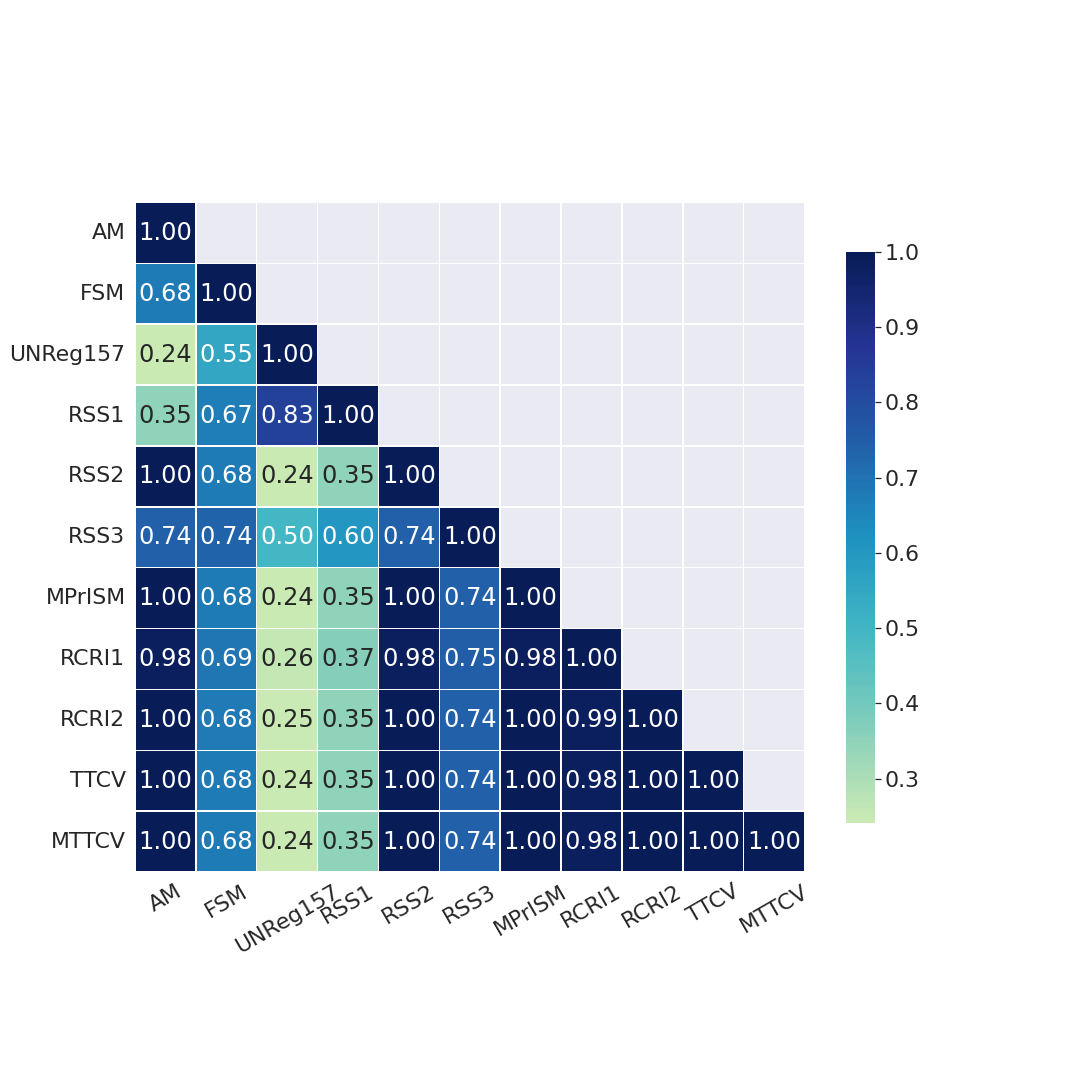}
  \caption{\footnotesize{Pairwise AIDs comparisons for metrics mapping $\mathcal{D}_{O_l}^H \rightarrow \mathbb{B}\times N_s$}.}
  \label{fig:highd_ol_b_times_t_hm}
\end{subfigure}%
\hspace{1em}%
\begin{subfigure}{0.30\linewidth}
  \centering
  \includegraphics[trim={0cm 4.5cm 5cm 5cm},clip,width=\textwidth]{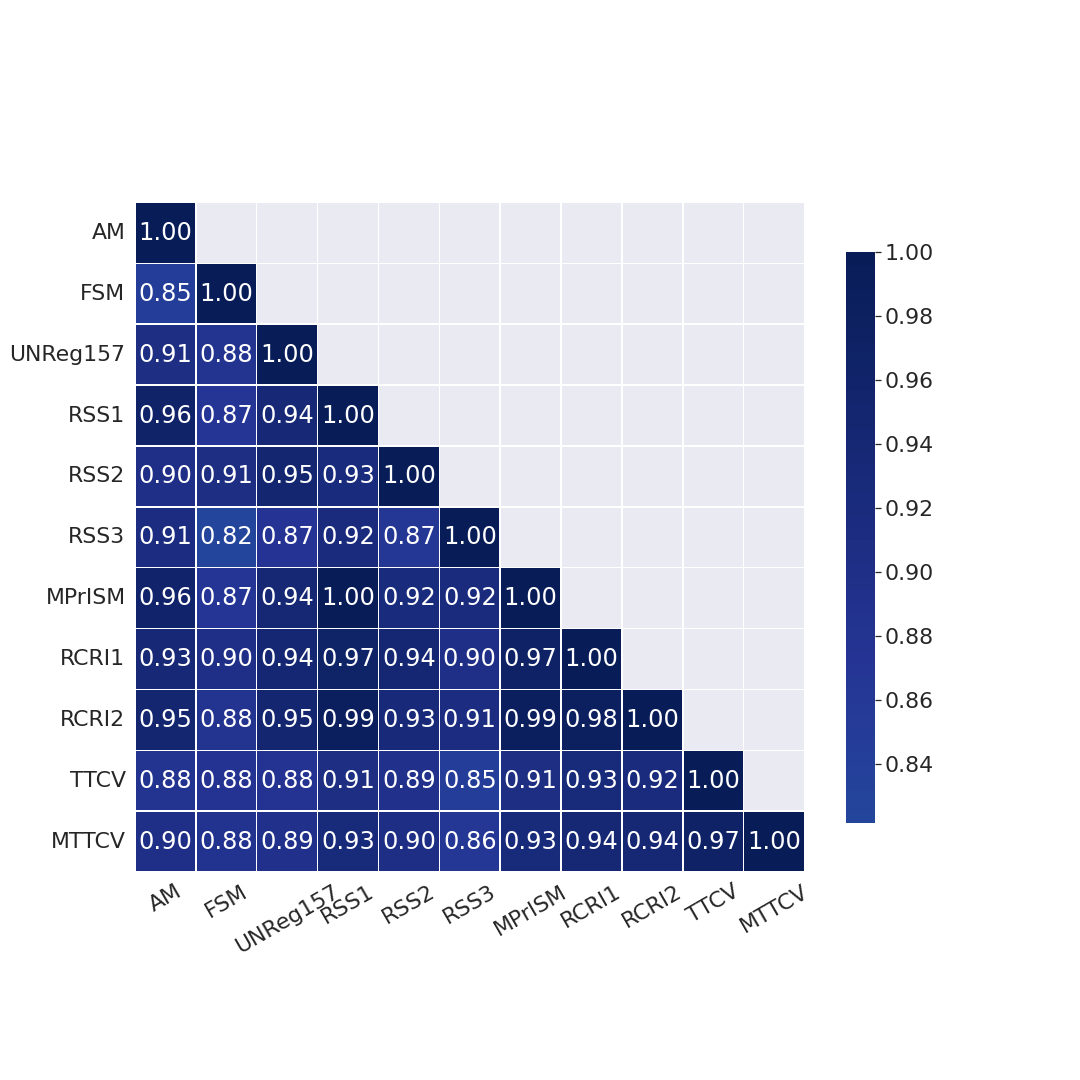}
  \caption{\footnotesize{Pairwise AIDs comparisons for metrics mapping $\mathcal{D}_{O_l}^V \rightarrow \mathbb{B}\times N_s$}.}
  \label{fig:vice_ol_b_times_t}
\end{subfigure}%
\hspace{1em}%
\begin{subfigure}{0.30\linewidth}
  \centering
  \includegraphics[trim={0cm 4.5cm 5cm 5cm},clip,width=\textwidth]{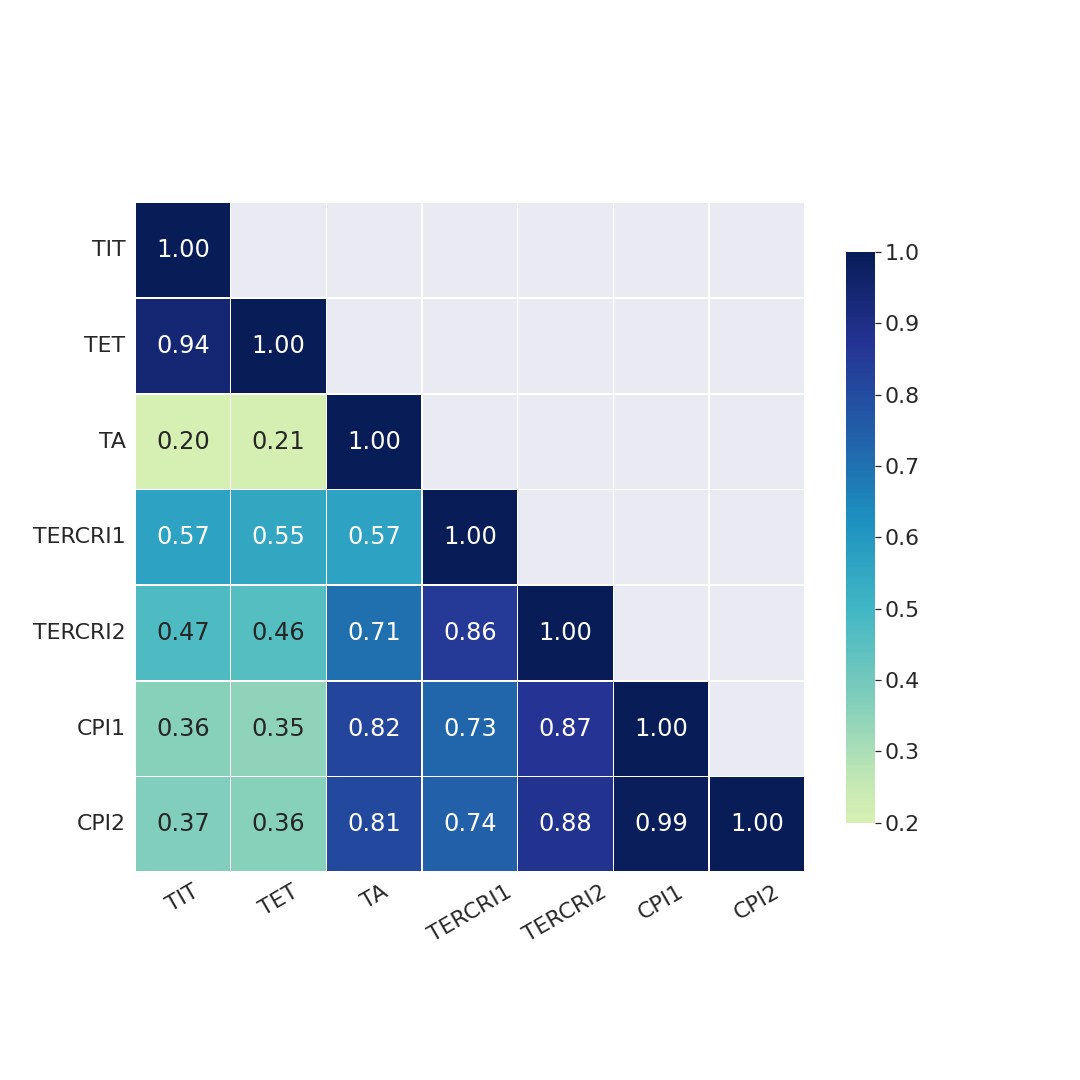}
  \caption{\footnotesize{Pairwise AIDs comparisons for metrics mapping $\mathcal{D}_{O_l}^V \rightarrow \mathbb{R}\times N_I$}.}
  \label{fig:vice_ol_xx_times_i}
\end{subfigure}
\newline
\begin{subfigure}{0.30\linewidth}
  \centering
  \includegraphics[trim={0cm 4.5cm 5cm 5cm},clip,width=\textwidth]{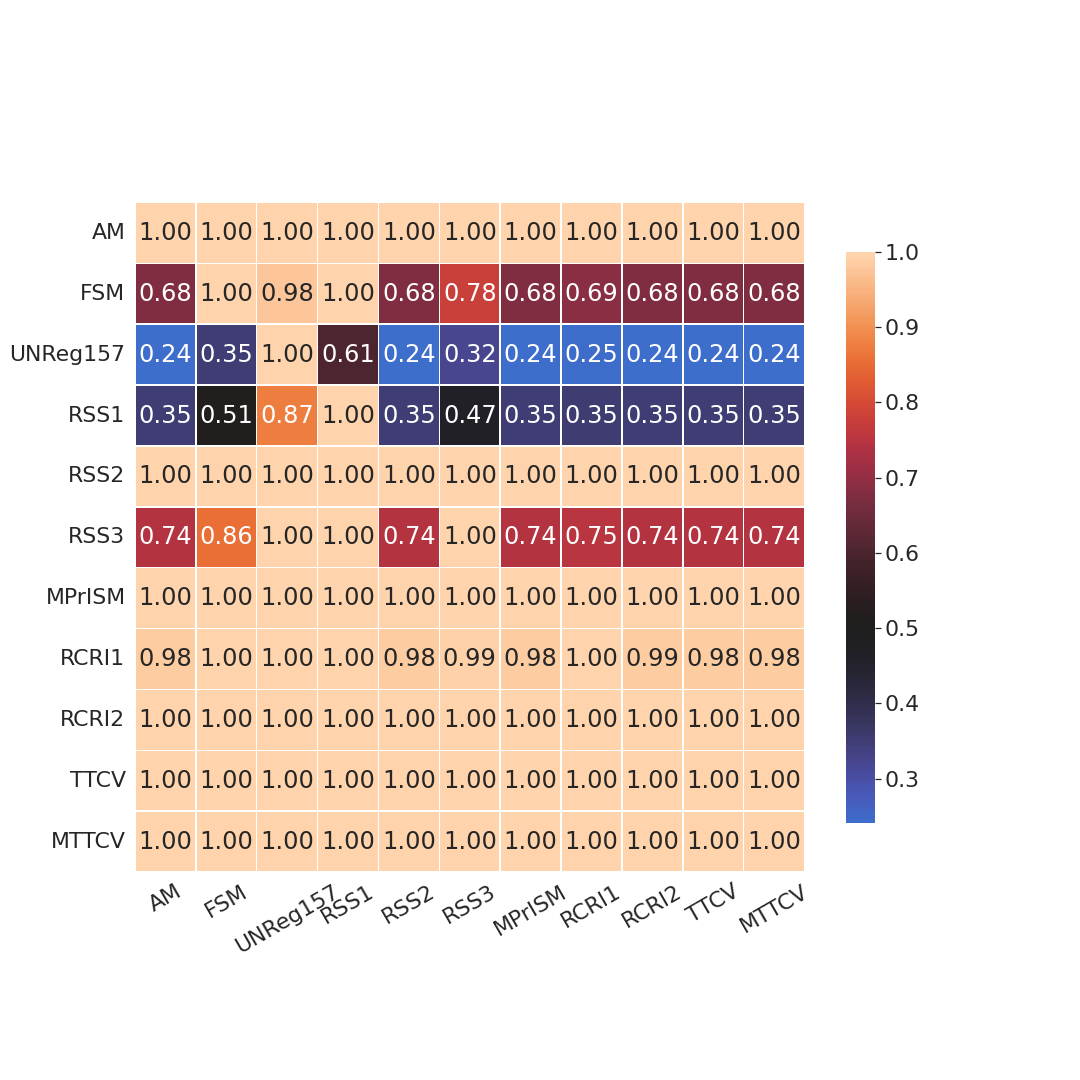}
  \caption{\footnotesize{Pairwise precision score comparisons for metrics mapping $\mathcal{D}_{O_l}^H \rightarrow \mathbb{B}\times N_s$}.}
  \label{fig:highd_ol_b_times_t_p_hm}
\end{subfigure}%
\hspace{1em}%
\begin{subfigure}{0.30\linewidth}
  \centering
  \includegraphics[trim={0cm 4.5cm 5cm 5cm},clip,width=\textwidth]{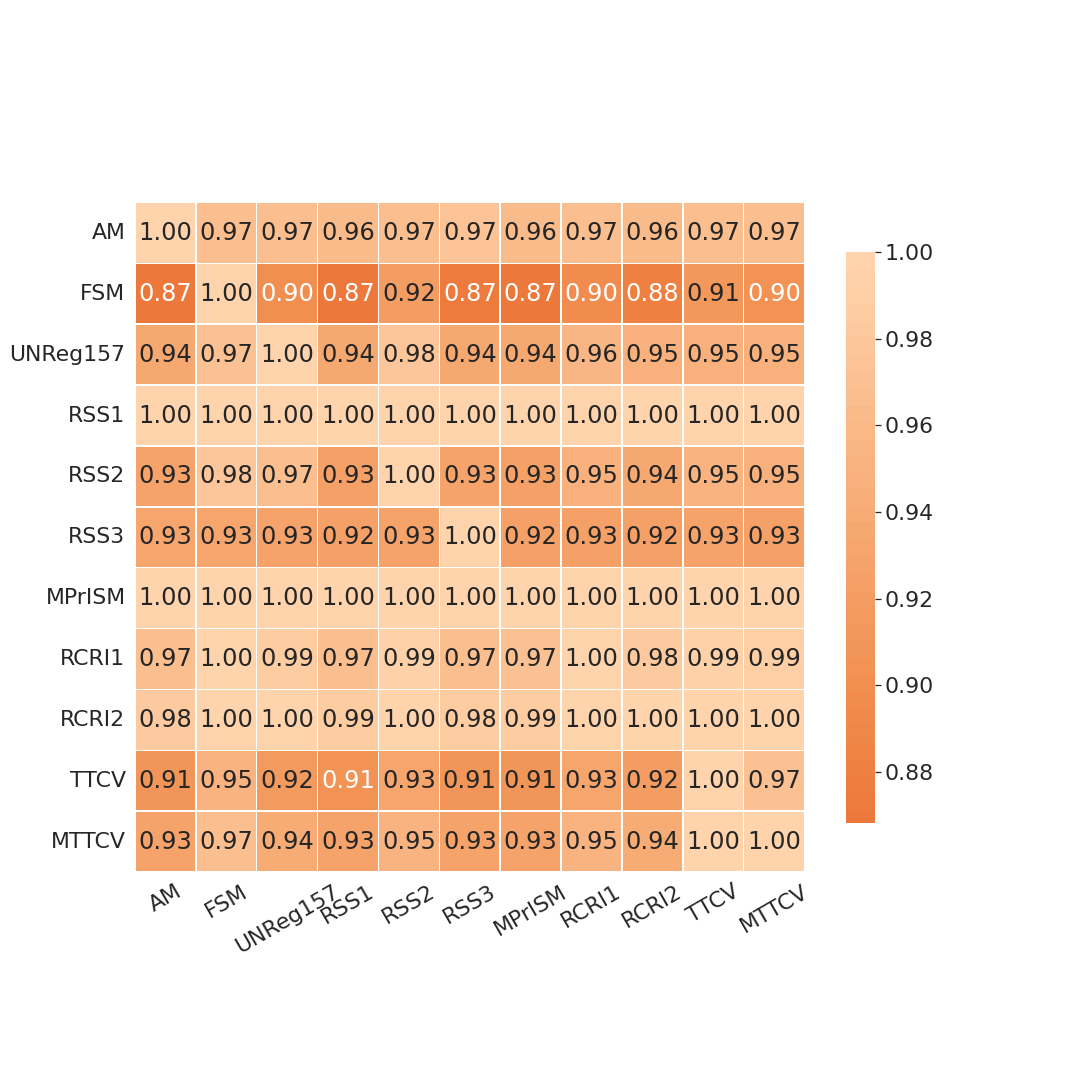}
  \caption{\footnotesize{Pairwise precision score comparisons for metrics mapping $\mathcal{D}_{O_l}^V \rightarrow \mathbb{B}\times N_s$}.}
  \label{fig:vice_ol_b_times_t_p_hm}
\end{subfigure}%
\hspace{1em}%
\begin{subfigure}{0.30\linewidth}
  \centering
  \includegraphics[trim={0cm 4.5cm 5cm 5cm},clip,width=\textwidth]{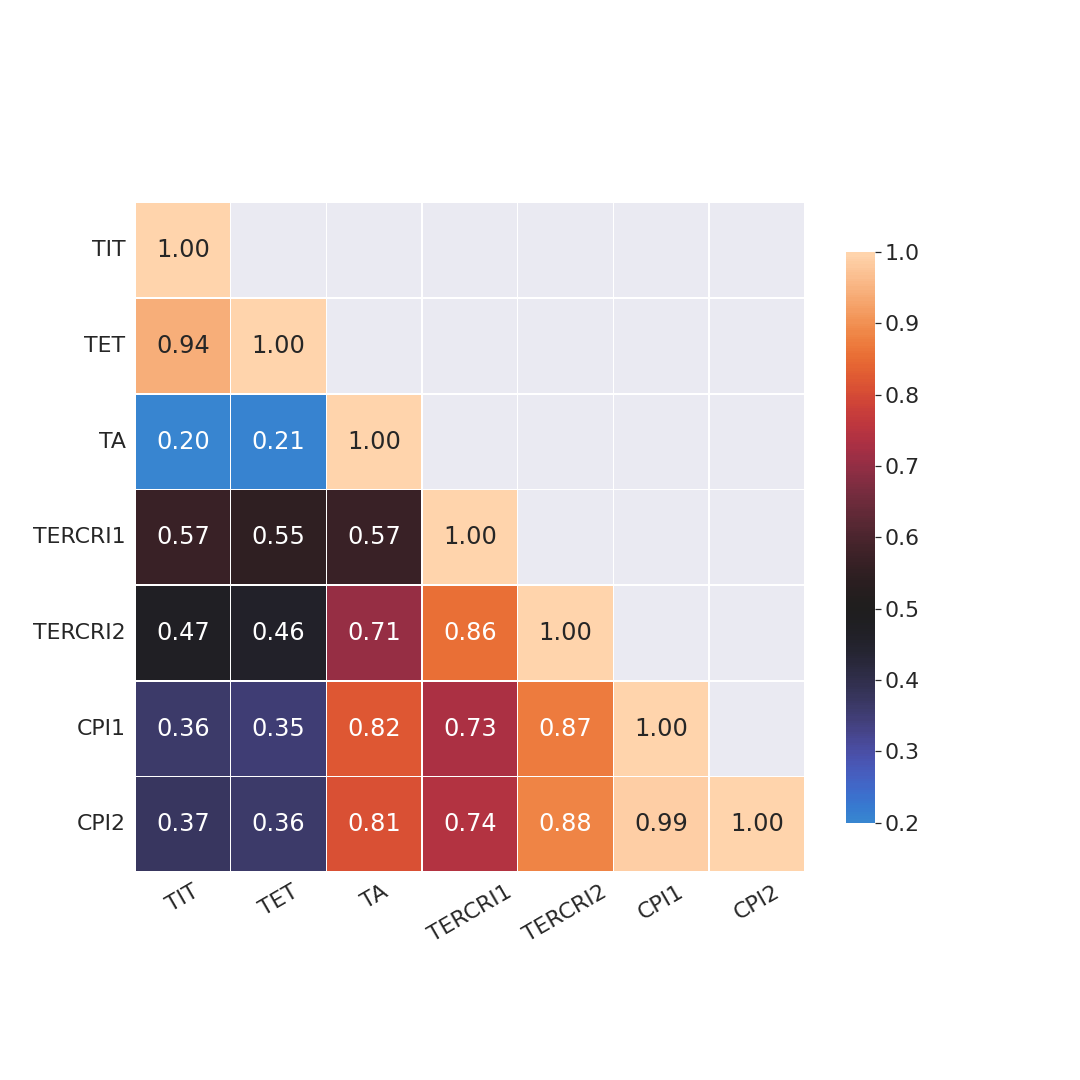}
  \caption{\footnotesize{Pairwise precision score comparisons for metrics mapping $\mathcal{D}_{O_l}^V \rightarrow \mathbb{R}\times N_I$}.}
  \label{fig:vice_ol_xx_times_i_mp_hm}
\end{subfigure}%
\newline
\caption{Pairwise comparisons applied to the data sets $\mathcal{D}_{O_l}^H$ and $\mathcal{D}_{O_l}^V$. }
\label{fig:ol_metrics_pairwise_agreement}
\vspace{-3mm}
\end{figure*}

Metric comparisons study performed using AID determination w.r.t. different metric output formats as explained in Section~\ref{sec:agreementjustification} and input data sets are presented in Fig.~\ref{fig:ol_metrics_pairwise_agreement}, showing the diversity in results obtained for the extracted data sets $\mathcal{D}_{O_l}^H$ and $\mathcal{D}_{O_l}^V$. The average AID (with standard deviation) for the first five AID demonstrating sub-figures in Fig.~\ref{fig:ol_metrics_pairwise_agreement} are 0.516($\pm$0.268), 0.663($\pm$0.198), 0.734($\pm$0.278), 0.924($\pm$0.044), and 0.666($\pm$0.262), respectively. Similar values for precision demonstrating sub-figures Fig.~\ref{fig:highd_ol_b_times_t_p_hm}, Fig.~\ref{fig:vice_ol_b_times_t_p_hm} and Fig.~\ref{fig:vice_ol_xx_times_i_mp_hm} are 0.666($\pm$0.263), 0.960($\pm$0.037), and 0.857($\pm$0.247), respectively. Note that the AID value of one in Fig.~\ref{fig:ol_metrics_pairwise_agreement} does not necessarily indicate the complete agreement. As the cardinalities for the input data sets are very large, any AID greater than 0.99 will show as one in the figure. For a similar reason, any AID smaller than 0.01 will show as zero. The same cause also justifies the illustrated values for the precision scores.

Diversity analysis results associated with $\mathcal{D}_{O_l}^H$ are shown in Fig.~\ref{fig:highd_ol_xx_times_t}, Fig.~\ref{fig:highd_ol_b_times_t_hm}, and Fig.~\ref{fig:highd_ol_b_times_t_p_hm}. A mixture of high and low values with significant variations highlight the performance differences among metrics. Moreover, the following observations are emphasized: (i) the extremely low AIDs and precision scores in Fig.~\ref{fig:highd_ol_b_times_t_hm} and Fig.~\ref{fig:highd_ol_b_times_t_p_hm} w.r.t. UNReg157 are due to significantly large number of states in the input data set with SV speeds above 60 km/h, which are deemed unsafe for all cases where a leading POV exists. This results in significantly low safe outcomes or True outcomes in UNReg157 compared to other metrics. (ii) RSS1 has lower values when comparing with the majority of other metrics as shown in Fig.~\ref{fig:highd_ol_b_times_t_hm} and Fig.~\ref{fig:highd_ol_b_times_t_p_hm} due to a large response time (1.924-second) for RSS1, which combined with high SV speeds, results in a significant over-estimation of unsafe states. (iii) Pairwise AIDs comparisons of MPrISM with other metrics are higher in general when considering comparison with metrics mapping to $\mathbb{B}\times N_s$ shown in Fig.~\ref{fig:highd_ol_b_times_t_hm} as compared to the metrics mapping to $\mathbb{R}\times N_s$ shown in Fig.~\ref{fig:highd_ol_xx_times_t}. This is because the classification, of all possible combined pairs of metric outcomes for extracted data set states, into three-class outcomes $\{-1,0,1\}$ emphasizes the subtle differences among the metric outputs. In the meanwhile, the direct use of Boolean classifications for Boolean output metrics may have hidden some of the details with the chosen threshold that differentiate the \texttt{True} outcomes from the \texttt{False} ones. (iv) Note MPrISM and RSS2 in Fig.~\ref{fig:highd_ol_b_times_t_hm} and Fig.~\ref{fig:highd_ol_b_times_t_p_hm} show almost same values. This is because within the ODD of $O_l$, the modeled and behavioral assumptions of MPrISM are almost equivalent to RSS without the response time assigned to the vehicles. Among the three RSS variants studied in this paper, RSS2 happens to have the smallest response time that is close to zero (0.117-second) as shown in TABLE~\ref{tab:rss_variants}.

Similar results associated with $\mathcal{D}^V_{0_l}$ are shown in Fig.~\ref{fig:vice_ol_xx_times_t}, Fig.~\ref{fig:vice_ol_b_times_t}, Fig.~\ref{fig:vice_ol_xx_times_i}, Fig.~\ref{fig:vice_ol_b_times_t_p_hm}, and Fig.~\ref{fig:vice_ol_xx_times_i_mp_hm}. As the VICE data set comprises of many safety-critical events (collisions and near-miss cases), the metrics tend to agree with each other more in this case as shown in Fig.~\ref{fig:vice_ol_b_times_t} and Fig.~\ref{fig:vice_ol_b_times_t_p_hm}.

In case of Fig.~\ref{fig:highd_ol_b_times_t_p_hm} and Fig.~\ref{fig:vice_ol_b_times_t_p_hm}, note that the values are not symmetric. This is due to different numbers of positive predictions between metrics as one has also revealed theoretically in~\eqref{eq:precision_c}. For example, with positives refering to the \texttt{True} classified outputs, FSM has 466641 positive outcomes while RSS3 has 512447 positive results. This leads to the precision scores of $\mathcal{P}_c(\text{FSM}, \text{RSS3}, \mathcal{D}^H_{O, k}) = 0.86$, whereas $\mathcal{P}_c(\text{RSS3}, \text{FSM}, \mathcal{D}^H_{O, k}) = 0.78$. Moreover, Fig.~\ref{fig:vice_ol_xx_times_i} and Fig.~\ref{fig:vice_ol_xx_times_i_mp_hm} show same results for metric outputs of the form of $\mathbb{R} \times N_I$. The precision scores in Fig.~\ref{fig:vice_ol_xx_times_i_mp_hm} are micro-averaged precision results, which aggregate the contributions of all three classes $\{-1, 0, 1\}$ in determining an average precision score. Hence, this becomes equivalent to AID for metric outputs of the form of $\mathbb{R} \times N_I$. For an example of disagreement between individual pairwise metrics, note $\mathcal{P}_c(\text{TIT}, \text{CPI2}, \mathcal{D}^H_{O, k})=0.37$, while $\mathcal{P}_c(\text{TIT}, \text{TET}, \mathcal{D}^H_{O, k})=0.94$. Note that the diversity can also be observed from the recall value analysis. For example, the recall value between FSM and UNReg157 w.r.t. $\mathcal{D}^H_{O, k}$ is 0.98 with a precision score of 0.35 (i.e., high-recall with low-precision). In the meanwhile, both the precision score and the recall value between FSM and RSS3 are relatively large (0.78 and 0.86). This reiterates the performance differences among metrics from the empirical perspective.

\begin{figure*}[!h]
\vspace{2mm}
\centering
\begin{subfigure}{0.30\linewidth}
  \centering
  \includegraphics[trim={0cm 0cm 0cm 0cm},clip,width=\textwidth]{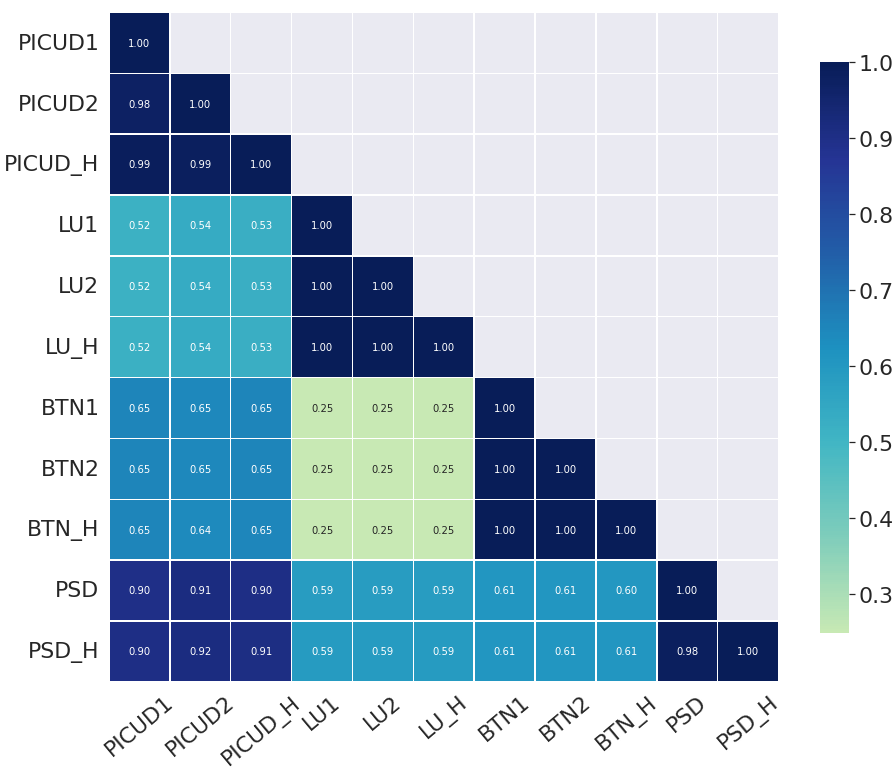}
  \caption{\footnotesize{Pairwise AIDs comparisons for metrics mapping $\mathcal{D}_{O_l}^H \rightarrow \mathbb{R}\times N_s$}.}
  \label{fig:highd_ol_r_times_t_hp_hm}
\end{subfigure}%
\hspace{1em}%
\begin{subfigure}{0.30\linewidth}
  \centering
  \includegraphics[trim={0cm 0cm 0cm 0cm},clip,width=\textwidth]{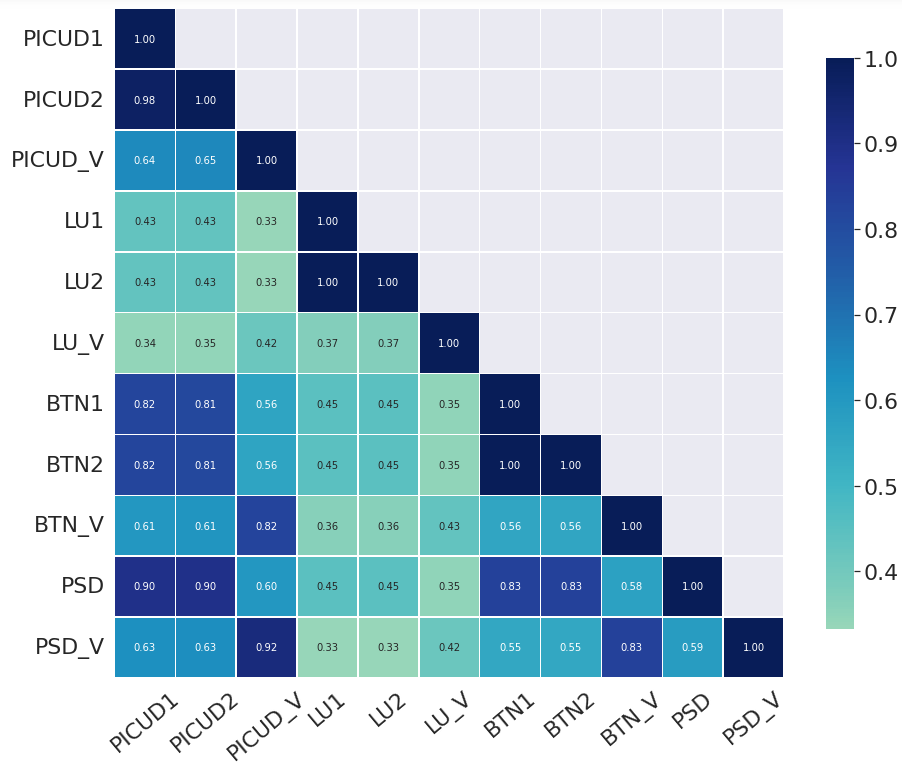}
  \caption{\footnotesize{Pairwise AIDs comparisons for metrics mapping $\mathcal{D}_{O_l}^V \rightarrow \mathbb{R}\times N_s$}.}
  \label{fig:vice_ol_r_times_t_hp_hm}
\end{subfigure}%
\hspace{1em}%
\begin{subfigure}{0.30\linewidth}
  \centering
  \includegraphics[trim={0cm 0cm 0cm 0cm},clip,width=\textwidth]{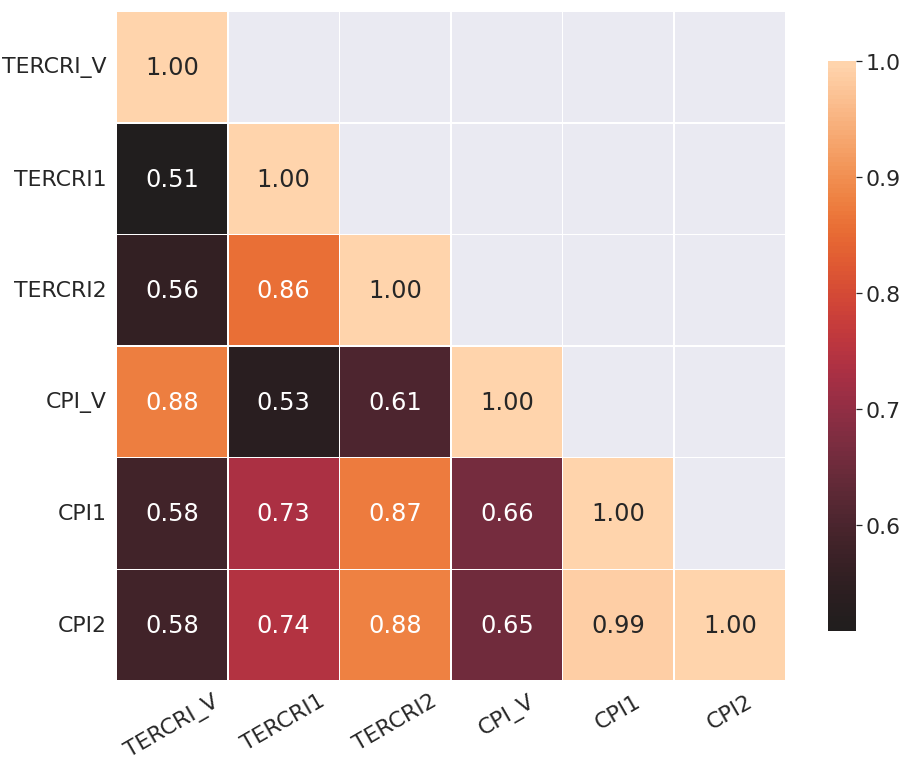}
  \caption{\footnotesize{Pairwise precision score comparisons for metrics mapping $\mathcal{D}_{O_l}^V \rightarrow \mathbb{R}\times N_I$}.}
  \label{fig:vice_ol_xx_times_i_mp_hp_hm}
\end{subfigure}%
\newline
\caption{Pairwise comparisons applied to the data sets $\mathcal{D}_{O_l}^H$ and $\mathcal{D}_{O_l}^V$ for studying effect of hyper-parameter values.}
\label{fig:ol_metrics_pairwise_agreement:hp_study}
\vspace{-3mm}
\end{figure*}

Additionally, Fig.~\ref{fig:ol_metrics_pairwise_agreement:hp_study} illustrates similar pairwise comparison results for metric variants involving hyper-parameters derived from the (i) used data sets (i.e., the HighD and the VICE), (ii) metric proposal references, and/or (iii) standard test procedures (explained earlier in Section~\ref{sec:metric_variants}). As observed in Fig.~\ref{fig:ol_metrics_pairwise_agreement}, a combination of high and low values demonstrate the disagreement between metrics. Note Fig.~\ref{fig:highd_ol_r_times_t_hp_hm} indicates a minor effect of hyper-parameter values on metric outcomes in case of the HighD data set, evident from similar values among variants of a metric. This is due to the large number of predominantly safe states in the HighD data set, which are not affected by change of hyper-parameter values among metric variants. Statistically, 90.5\% of states in HighD have DHW greater than 20-metre, and only 10.9\% of states have the SV approaching leading POV with greater than 2~$\mathrm{m/s}$ approach speed. On the other hand, Fig.~\ref{fig:vice_ol_r_times_t_hp_hm} and Fig.~\ref{fig:vice_ol_xx_times_i_mp_hp_hm} illustrate a significant effect of hyper-parameter values on metric outcomes for the VICE data set, evident by the differences in results between data set based metric variants and other generalised metric variants. 

\begin{figure*}[!h]
    \centering
    \includegraphics[width=0.98\textwidth]{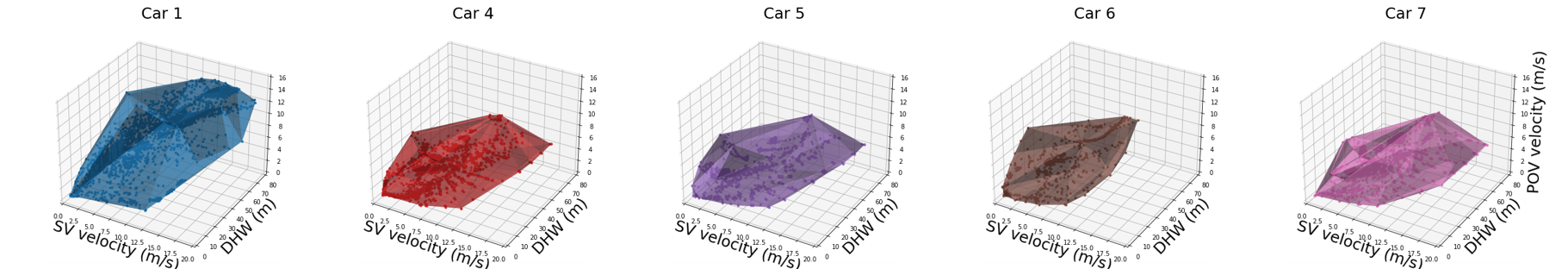}
    \caption{The $\bar{\epsilon}\alpha$-almost safe sets obtained from $\{\mathcal{D}^V_{O_l, i}\}_{i\in\{1,4,5,6,7\}}$.}
    \label{fig:vice_ass}
\end{figure*}

Moreover, CR, FMRI, and $\bar{\epsilon}\alpha$-ASS are characterizations of the complete input data set without specifications for each individual state or incident. Their results for some of the SVs in the VICE data set (more specifically $\{\mathcal{D}^V_{O_l, i}\}_{i\in\{1,4,5,6,7\}}$) are shown in Fig.~\ref{fig:vice_ass} and TABLE~\ref{tab:vice_comp}. In Fig.~\ref{fig:vice_ass}, $\bar{\epsilon}\alpha$-ASS results are illustrated in the minimally defined three-dimensional state space for $O_l$, comprising of the three most important features as explained earlier in Section~\ref{sec:preliminaries}. The $\bar{\epsilon}\alpha$-almost safe sets are shown and the points inside the sets are the states in the extracted data sets. The expected probability for the SV to remain safe (i.e., inside the illustrated set) is revealed by $1-\bar{\epsilon}$. Smaller the value of $\bar{\epsilon}$, lower the risk. The occupancy is an indicative feature of the size of the almost safe set. The higher the value, the more the state space covered by the states in the data set. The density compares the number of actual states inside the safe-set to the size of the safe-set. 
A statistical summary w.r.t. the individual SV in VICE is also included in TABLE~\ref{tab:vice_comp} for some of the metrics. Car 2 and Car 3 are ignored for this study as the cardinalities of the respective extracted data sets are too small. It is also not too surprising to see that the metrics form a diverged opinion regarding the least well-performed SV in the VICE data set. Note that FMRI requires consecutive observation of safe driving mileage, hence it is not applicable for input data sets of unsafe events. This explains the FMRI value of one for Car 5, 6, and 7 in TABLE~\ref{tab:vice_comp}. It is also noted here that the values for RSS2 and RSS3 appear same due to the large cardinality of the data set as stated earlier in this section. 

\begin{table}[!b]
    \caption{Statistical inference metric outcomes and the statistical summary (the occurrence rate of Boolean outcomes related to high-risk and unsafe states) of metrics applied to the VICE data set, as shown in Fig.~\ref{fig:vice_ol_b_times_t}. The SV deemed of the most negative performance (i.e., the most unsafe vehicle or the one of the highest risk) is highlighted in bold font.}
    \label{tab:vice_comp}
    % \vspace{2mm}
    \centering
    \resizebox{0.49\textwidth}{!}{%
    \begin{tabular}{l|l|l|l|l|l|l}
\hline
Metric                                      & Property                            & Car 1  & Car 4 & Car  5 & Car 6  & Car 7 \\ \hline
CR                                          & rate ($10^{-5}$)                    & 0      & 0     & 1.99   & 0.31   & \textbf{2.87}  \\ \hline
FMRI                                          & rate ($10^{-2}$) & 14     & 35    & \textbf{100}    & \textbf{100}    & \textbf{100}   \\ \hline
\multirow{3}{*}{$\bar{\epsilon}\alpha$-ASS} & $\bar{\epsilon}$ ($10^{-5}$)        & 277    & 395   & 809    & \textbf{1636}   & 545   \\ \cline{2-7} 
                                            & occupancy                           & 0.296  & 0.149 & 0.151  & \textbf{0.097}  & 0.151 \\ \cline{2-7} 
                                            & density                             & 1.259  & 1.752 & 0.906  & \textbf{0.698}  & 1.346 \\ \hline
FSM                                         & \multirow{10}{*}{unsafe state rate} & 0.060  & 0.142 & \textbf{0.302}  & 0.075  & 0.286 \\ \cline{1-1} \cline{3-7} 
UNReg157                                    &                                     & 0.057  & 0.044 & \textbf{0.142}  & 0.013  & 0.132 \\ \cline{1-1} \cline{3-7} 
RSS1                                        &                                     & 0.0    & 0.003 & 0.006  & 0.0    & \textbf{0.009} \\ \cline{1-1} \cline{3-7} 
RSS2                                        &                                     & 0.057  & 0.065 & 0.148  & 0.029  & \textbf{0.164} \\ \cline{1-1} \cline{3-7} 
RSS3                                        &                                     & 0.057  & 0.065 & 0.148  & 0.029  & \textbf{0.164} \\ \cline{1-1} \cline{3-7} 
MPrISM                                      &                                     & 0.0001 & 0.002 & 0.011  & 0.0008 & \textbf{0.012} \\ \cline{1-1} \cline{3-7} 
RCRI1                                       &                                     & 0.004  & 0.039 & \textbf{0.109}  & 0.015  & 0.082 \\ \cline{1-1} \cline{3-7} 
RCRI2                                       &                                     & 0.0004 & 0.013 & \textbf{0.065}  & 0.006  & 0.043 \\ \cline{1-1} \cline{3-7} 
TTCV                                        &                                     & 0.038  & 0.153 & \textbf{0.203}  & 0.054  & 0.165 \\ \cline{1-1} \cline{3-7} 
MTTCV                                       &                                     & 0.033  & 0.118 & \textbf{0.157}  & 0.046  & 0.111 \\ \hline
\end{tabular}%
    }
    % \vspace{-5mm}
\end{table}

\begin{figure}[!h]
    \centering
    \includegraphics[trim={1cm 1cm 1cm 3cm},clip,width=0.48\textwidth]{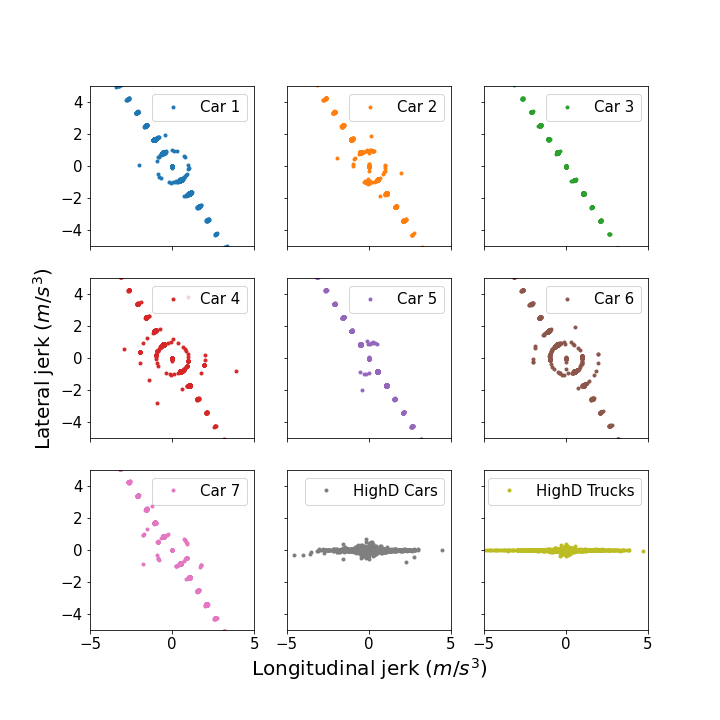}
    \caption{Jerk results for $\mathcal{D}^V_{O_0}$ and $\mathcal{D}^{H}_{O_0}$.}
    \label{fig:jerk}
\end{figure}

Finally, the Jerk results w.r.t. $\mathcal{D}^V_{O_0, i}, i\in\Z_7$, $\mathcal{D}^H_{O_0, \text{car}}$, and $\mathcal{D}^H_{O_0, \text{truck}}$ are shown in Fig.~\ref{fig:jerk}. Note that Jerk is a special metric that only relies on the SV states, hence the SV Oriented ODD, $O_0$, is adopted. Recall that vehicles in the VICE data set are more frequently exposed to safety-critical events. This explains the higher magnitude of jerks in comparison with the HighD results. Moreover, as the VICE data set was extracted from standard ADAS testing scenarios, the vehicles always make the same type of single lane change, hence the large-magnitude jerks exhibit the particular pattern shown in Fig.~\ref{fig:jerk}. 

\section{Conclusions and Discussions} \label{sec:conclusion}
This section summarizes the paper with some added discussions of related literature and future work of interest.

\subsection{Other Related Literature}\label{sec:oliterature}
Note that a metric, in its original form, is not necessarily created exclusively for the prescribed working purpose in a post-processing fashion. For example, THW, RLA, and BTN~\cite{Jansson2005CollisionAT} were proposed as an online driving risk indicator, RSS~\cite{shalev2017formal} was intended as a safe control action supervisor that can be added to any driving decision-making module, FSM~\cite{mattas2020fuzzy} was defined to justify if a particular driving scene would evolve to an unpreventable collision, to name a few. Some of the other works involve formal verification methods~\cite{mitra2021verifying, meng2022learning} such as those using set-based approaches~\cite{althoff2014online, klischat2019generating} to determine reachable states under different kinds of input assumptions (e.g., uncertainties, steady-state, instantaneous control action, etc.). These primarily involve behavioral models akin to the metrics considered in this work, and they are inherently aimed to complement the decision-making process by acting as supervisory controllers. Hence they have not been included in the diversity analysis as the online performance verification is not the focus of this study.

Moreover, some other metrics are not involved in the diversity analysis for various reasons. First, criticality metric~\cite{junietz2018criticality}, Safety Force Field (SFF) method~\cite{nvidia2019}, Enhanced Time-to-Collision (ETTC)~\cite{feng2020testing, chen2016comparison}, Minimum Safe Distance Violation (MSDV)~\cite{wishart2020driving}, Minimum Safe Distance Calculation Error (MSDCE)~\cite{wishart2020driving}, Proper Response Action (PRA)~\cite{wishart2020driving}, and Steer Threat Number (STN)~\cite{Jansson2005CollisionAT, eidehall2011multi} are not studied, as they all have close analogous among the 33 studied metrics~\cite{suk2022rationale}. Second, (Post) Encroachment Time~\cite{allen1978analysis} ((P)ET), Predictive Encroachment Time (PrET)~\cite{allen1978analysis}, T2~\cite{laureshyn2010evaluation}, and Time Advantage (TAdv)~\cite{hansson1975studies} are not studied, as they all require a \emph{conflict point}, defined as the single-point intersection between the motion trajectories of the SV and the POV. The existence of such a conflict point is not always valid in the studied data sets. Moreover, Delta-V~\cite{shelby2011delta}, Extended Delta-V Indicator~\cite{laureshyn2010evaluation}, Conflict Severity~\cite{bagdadi2013estimation}, and Conflict Index~\cite{alhajyaseen2015integration} not only require the existence of the conflict point, but also rely on the vehicles' mass information, which is not available in some of the studied data sets. Some of the metrics like TTX (e.g., Time to Brake (TTB), Time to Steer (TTS), Time to Kickdown (TTK), and Time to React (TTR))~\cite{hillenbrand2006multilevel, tamke2011flexible} involve behavior models (e.g., instantaneous acceleration for the POV) similar to the metrics considered in this study. Assuming these behavior models, the metrics determine the maximum remaining time for the SV to begin a collision avoidance maneuver. This can be maximum deceleration, maximum acceleration or maximum steering in either direction. Due to similarity in behavior with acceleration based metrics studied in this paper, and the lack of parameter information in the studied data sets related to vehicle geometry, minimum turning radius, these metrics have not been included in the study. Worst Time to Collision (WTTC)~\cite{wachenfeld2016worst} and Instantaneous Safety Metric (ISM)~\cite{barickman2019instantaneous} assume a constant set of control actions throughout the predictive horizon. They are not included in the study, as the worst case situation is better taken into consideration by MPrISM, which performs optimization across consecutive time steps over a pre-defined time horizon, to determine the shortest time leading to a collision. From the different categories of metrics described in~\cite{dahl2018collision}, the diversity analysis covers single behavior metrics (e.g., TTC, PSD, and DRAC), optimization based methods (e.g., MPrISM), and probabilistic methods (e.g., FMRI and $\bar{\epsilon}\alpha$-Almost Safe Set). Accepted Gap Size (AGS)~\cite{petzoldt2014relationship}, Space Occupancy Index (SOI)~\cite{ogawa2007analysis}, and Pedestrian Risk Index (PRN)~\cite{cafiso2011crosswalk} are not incorporated, as they are designed for ODDs that are mostly pedestrian-related and are not of primary interest of this paper. Finally, the selected state features in the HighD and VICE data sets have also limited the scope of our study, as some metrics require unavailable information, such as Collision Incident (CI)~\cite{wishart2020driving}, ADS active (ADSA)~\cite{wishart2020driving}, Rules-of-the-Road Violation (RRV)~\cite{wishart2020driving}, Achieved Behavioral Competency (ABC)~\cite{wishart2020driving}, Human Traffic Control Detection Error Rate (HTCDER)~\cite{wishart2020driving}, Human Traffic Control Violation Rate (HTCVR)~\cite{wishart2020driving}, and Time to Zebra (TTZ)~\cite{varhelyi1998drivers}. 

The selected metrics in this paper cover a broad spectrum of various related works in the literature. One can refer to other summarized reports and surveys on this topic~\cite{griffor2019workshop,wishart2020driving,wang2021review,westhofen2022criticality,elli2021evaluation,jammula2022evaluation,ieee2022white,kidambi2022sensitivity, como2023evaluating} for other metrics in the literature. Note that this paper focuses on applying metrics to ADAS or ADS specific applications. However, the metric applicability is fundamentally independent of the subject that drives or controls the vehicle. As a result, the metrics discussed in this paper are also applicable to conventional vehicles driven by human drivers.

\subsection{Conclusion}
The work presented in this paper, provides an analytical and empirical understanding of the distinctions and correlations among the majority of the widely used or studied safety metrics for performance assessment of human-driven and ADS equipped vehicles. The metrics are reviewed theoretically in terms of their construction to understand their input and output, and if the assumptions behind their formulation are reasonable for practical use. Based on the analytical study, three types of categorical algorithms are proposed that cover all metrics reviewed in this paper and serve as a metric creation recipe. This paper also shows that empirical justification cannot serve as a rationale for metric acceptability by conducting an empirical analysis for the lead-vehicle interaction ODD of a naturalistic data set (HighD) and a high-risk data set (VICE). An overall agreement index of 0.650 $\pm 0.262$ and precision of 0.867 $\pm 0.218$ is obtained, highlighting that the metrics do not show a dominant concurrence. Some other metrics also fail to achieve an agreement on the best-performed vehicle revealed by the VICE data set.

The authors propose this analysis approach to help bridge the gap in understanding the theoretical nature and the applications of metrics. As part of the future work, this research can be expanded to include larger and more diverse data sets with more vehicles and more ODDs. 

\appendices
\section{}\label{apx:notation}

\nomenclature[01]{$\mathcal{S}$}{Set of observed states}
\nomenclature[02]{$\s$}{A state in set $\mathcal{S}$}
\nomenclature[03]{$O$}{Operational Design Domain (ODD)}
\nomenclature[04]{$\mathcal{D}$}{Set of time-dependent states}
\nomenclature[05]{$T$}{Set of time-stamps}
\nomenclature[06]{$t$}{A time-stamp in set $T$}
\nomenclature[07]{$\mathcal{I}$}{Set of states consecutive in time sharing the same SV}
\nomenclature[08]{$O_0$}{SV oriented ODD}
\nomenclature[09]{$O_l$}{Lead-vehicle interaction ODD}
\nomenclature[10]{$\mathcal{D}^H$}{Extracted HighD data set}
\nomenclature[11]{$\mathcal{D}^V$}{Extracted VICE data set}
\nomenclature[12]{$\mathcal{D}_{O_l}$}{Set of states from the data set $\mathcal{D}$ occurred within ODD $O_l$}
\nomenclature[13]{$\mathcal{D}_{O,k}$}{Set of states from the data set $\mathcal{D}$ occurred within ODD $O$ and associated with selected SV represented by an index $k$}
\nomenclature[14]{$\mathcal{D}_{O_0,k}$}{Set of states from the data set $\mathcal{D}$ occurred within ODD $O_0$ and associated with selected SV represented by an index $k$}
\nomenclature[15]{$\mathcal{D}_{O_l,k}$}{Set of states from the data set $\mathcal{D}$ occurred within ODD $O_l$ and associated with selected SV represented by an index $k$}
\nomenclature[16]{$\mathcal{M}$}{A vehicle safety performance metric}
\nomenclature[17]{$G$}{Set of metric outcomes}
\nomenclature[18]{$g$}{A metric outcome in set $G$}
\nomenclature[19]{$N_s$}{The number of states in $\mathcal{D}_{O_0,k}$}
\nomenclature[20]{$N_I$}{The number of incidents in $\mathcal{D}_{O_0,k}$}
\nomenclature[21]{$f_0$}{A motion model for the SV}
\nomenclature[22]{$f_1$}{A motion model for the POV}
\nomenclature[23]{$b_0$}{A behavior model for the SV}
\nomenclature[24]{$b_1$}{A behavior model for the POV}
\nomenclature[25]{$\tau$}{Length of an incident}
\nomenclature[26]{$R$}{Set of metric outcomes}
\nomenclature[27]{$\mathcal{D}^H_{O_l}$}{Extracted data set from the HighD data set}
\nomenclature[28]{$\mathcal{D}^V_{O_l}$}{Extracted data set from the VICE data set}
\nomenclature[29]{$\bar{\mathcal{A}}_{ij}^{m}$}{Three-class classification outcome in $\{-1, 0, 1\}$ for comparing two metrics $\mathcal{M}_m$ ($m \in \{1,2\}$) in terms of safety performance of states $\s_i$ and $\s_j$}
\nomenclature[30]{$\mathcal{C}^m_{O, k}$}{Set of classification outcomes transformed through using metric $\mathcal{M}_m$ on extracted data set $\mathcal{D}_{O, k}$}
\nomenclature[31]{$\mathcal{A}$}{Agreement Index (AID)}
\nomenclature[32]{$\mathcal{P}_c$}{Precision score}
\nomenclature[33]{$\bar{\epsilon}$}{The probability coefficient for the almost safe set}
\printnomenclature

\section{Individual Metric Overview}\label{apx:constructive_da:overview} 

\textbf{Time to Collision (TTC)}, in its original form~\cite{hayward1972near}, measures the time the SV takes to collide with a leading POV, assuming both vehicles are in steady-state on straight road segments. In this paper, the particular formulation for TTC is defined through a ratio as $\frac{\text{dhw}}{dv}$, where dhw and $dv$ denote the bumper-to-bumper distance headway and the relative velocity between the two vehicles in a car-following scenario, respectively. The relative velocity is defined as the difference between the SV velocity and the leading POV velocity.

\textbf{Potential Time to Collision (PTTC)}~\cite{wakabayashi2003traffic} is a modified form of TTC metric which measures the time the SV takes to collide with a leading POV, assuming the SV maintains steady-state behavior and the POV follows current instantaneous control (deceleration only). 

\textbf{Modified Time to Collision (MTTC)}~\cite{Ozbay2008Derivation} is another modified form of the TTC metric which measures the time the SV takes to collide with a leading POV, but follows the assumption that both the SV and the POV maintain their instantaneous control action (acceleration value) and current heading angle indefinitely in a car-following scenario.

\textbf{Model Predictive Instantaneous Safety Metric (MPrISM)}~\cite{weng2020model} determines the risk of the SV by taking into account the worst-case maneuver for the POV and the optimal collision avoidance strategy for the SV. It further determines the SV's proximity to collision through a predicted TTC within a fixed horizon (one-second in this paper).

\textbf{Potential Index for Collision with Urgent Deceleration (PICUD)}~\cite{uno2002microscopic} measures the distance between the SV and a leading POV assuming (i) the SV maintains the instantaneous speed during a constant reaction time and brakes-to-stop afterwards, and (ii) the POV brakes-to-stop.

\textbf{Difference of Space Distance and Stopping Distance (DSS)}~\cite{okamura2011impact} is fundamentally equivalent to PICUD with only minor differences regarding the choices of hyper-parameters.

\textbf{Deceleration Rate to Avoid the Crash (DRAC)}~\cite{cooper1976traffic} defines the SV's instantaneous longitudinal control action required for collision avoidance against a leading POV assuming steady-state POV behavior.

\textbf{Deceleration to Safety Time (DST)}~\cite{hupfer1997deceleration} defines the SV's required deceleration to maintain a desired safe time gap w.r.t. the leading POV, assuming steady-state POV behavior.

\textbf{Required Longitudinal Acceleration (RLA)}~\cite{Jansson2005CollisionAT} defines the SV's longitudinal acceleration required to ensure a zero relative velocity at the time of impact, assuming instantaneous control of the POV.

\textbf{Reciprocal Time to Collision (RTTC)}~\cite{chin1992quantitative} defines the notion of risk as the inverse of TTC and intuitively, denotes a conflict severity measure which increases with the increase of risk as opposed to TTC, which decreases with the increase of risk. 

\textbf{Brake Threat Number (BTN)}~\cite{Jansson2005CollisionAT} is defined at each state $s \in {D}_{0,k}$ as the ratio of RLA and the maximum allowed deceleration of the SV.

\textbf{Time to Accident (TA)}~\cite{perkins1967traffic} is defined as the TTC value for the state $\s(t)$ when a SV evasive action is initially identified. Evasive action is defined in~\cite{dingus2006100} as any maneuver which exceeds the longitudinal acceleration of 4.95 $\mathrm{m/s^2}$ or the lateral acceleration of 3.92 $\mathrm{m/s^2}$.  

\textbf{Crash Potential Index (CPI)}~\cite{cunto2008assessing} is defined as the ratio of time duration in an incident when the SV cannot avoid a crash (determined by comparing DRAC value to the maximum allowed deceleration) to the total time duration of the incident.

\textbf{Time Exposed TTC (TET)}~\cite{minderhoud2001extended} is an incident focused metric which measures the duration of time in an incident when TTC is below a set threshold TTC value.

\textbf{Time Integrated TTC (TIT)}~\cite{minderhoud2001extended} is also an incident based metric similar to TET and measures the duration as well as the extent to which TTC is lower than a set threshold TTC value, by calculating the sum of areas between the TTC curve and the set threshold curve at each time instant when TTC is less than the threshold value.   

\textbf{Rear End Collision Risk Index (RCRI)}~\cite{oh2006method} compares the stopping distance of the SV and a leading POV in a car-following scenario, assuming both vehicles brake to stop with the maximum allowed deceleration. The brake-to-stop action for the SV takes into account an initial time delay, during which the speed is maintained constant. If the distance covered by the SV, during the time delay and the time taken to come to stop after that, is greater than the sum of the initial headway distance and the POV stopping distance, the SV will collide with the POV before coming to a stop. Hence, RCRI outputs a Boolean \texttt{False} in such a case, deeming the SV as unsafe.

\textbf{Time Exposed Rear End Collision Risk Index (TERCRI)}~\cite{rahman2018longitudinal} is based on RCRI and calculates the duration of time in an incident when the SV is deemed unsafe as defined by RCRI. 

\textbf{Time to Collision Violation (TTCV)}~\cite{mahmud2017application} is a modification of TTC metric. Given a threshold TTC value, TTCV gives a Boolean output of \texttt{False} if the calculated TTC value is below the set threshold and \texttt{True} otherwise.

\textbf{Modified Time to Collision Violation (MTTCV)}~\cite{Ozbay2008Derivation} is a modification of MTTC, similar to TTC-to-TTCV modification. 

\textbf{Responsibility Sensitive Safety (RSS)}~\cite{shalev2017formal} conceptually formalizes the human understanding of traffic laws and common sense driving rules to ensure that the AV would not be the cause of an accident and would reach a minimum risk state in case of unavoidable accidents. RSS was devised to be used along with a decision-making controller, supervising over the actions based on formalized rules. In order to be used as a safety metric, the idea of RSS is adapted to get the longitudinal and the lateral safe distances for each state $\s$, which are used as recommended thresholds. The SV is deemed safe if either of the thresholds is respected.

\textbf{Fuzzy Safety Model (FSM)}~\cite{mattas2020fuzzy} takes defensive driving by humans into account (comfortable deceleration at an early stage) in anticipation of emergency situations to avoid extreme harsh actions later. It comprises of lateral and longitudinal safety checks performed in order to calculate a deceleration reaction by the SV using fuzzy logic. In the specific implementation of FSM performed in this paper, only the safety checks are performed to obtain a Boolean output of \texttt{True} for the safe situation and \texttt{False} otherwise.

\textbf{UN Regulation 157 (UNReg157)}~\cite{united2020proposal} is a type approval requirement for automated lane keeping system (ALKS) operating in a constrained ODD (below the speed of 60 km/h on motorways with no lane changes possible). The performance requirements under this regulation are adopted as safety checks for the SV in this paper. Section 5.2.3.3 in~\cite{united2020proposal} gives the minimum following distance based on the SV speed for a car-following scenario. An actual following distance less than the minimum suggested distance deems the SV as unsafe (leading to an unpreventable collision). For the case of cut-in scenarios, Section 5.2.5.2 recommends a threshold TTC value. If the actual TTC is less than the threshold, the SV is deemed unsafe. UNReg157 is recommended for the SV speeds below 60 km/h. For the purpose of compatibility with all states $s \in \mathcal{D}_{O,k}$, the SV is deemed safe above 60 km/h only if no leading POV exists.

\textbf{Crash Index (CI)}~\cite{Ozbay2008Derivation} is a modification of MTTC. CI takes into account the severity as well as the likelihood of a potential conflict between the vehicles in a car-following scenario, expressed in the form of effect of speed on the kinetic energy involved in collision and the elapsed time before collision, respectively. 

\textbf{Proportion of Stopping Distance (PSD)}~\cite{astarita2019traffic} is defined as a ratio determined for each state $s \in {D}_{0,k}$. The denominator defines the SV's distance to stop assuming the maximum allowed deceleration. The numerator expresses the distance headway between the SV and the POV.

\textbf{Aggressive Driving (AD)} gives a Boolean output for each defined incident in the given input data set $\mathcal{D}_{O, k}$. The output is a \texttt{True} value if there are at least three violations of acceleration limits (either longitudinal or lateral or both) by the SV within a 60-second time duration, with each violation lasting a minimum of one-second (all mentioned parameters are from~\cite{wishart2020driving}).

\textbf{Accident Metric (AM)}, in its original form~\cite{otte2003scientific}, shares the same output format with AD as $\mathbb{B} \times N_I$. The metric returns \texttt{True} for every incident that terminates with a collision event and \texttt{False} otherwise. In this paper, another variant of AM is also considered by identifying at each state $s \in \mathcal{D}_{O, k}$ whether a SV-involved collision event has taken place or will take place, leading to the output space of $\mathbb{B} \times N_s$. 

\textbf{Jerk}~\cite{bellem2018comfort} refers to the rate of change of acceleration, longitudinally and laterally, with respect to time. The metric forms a mapping $\mathcal{D}_{O, k} \rightarrow \R^2 \times N_s$. Note that the jerk measure is normally considered as a comfort measure~\cite{bellem2018comfort}. However, higher jerk value is occasionally considered as an indicator of a greater risk~\cite{bagdadi2013development}, which can be adapted to safety performance justification purposes.

\textbf{Gap Time (GT)}~\cite{vogel2002characterizes} is defined as the exact time duration for a reference point on the SV to reach a reference line defined by the POV. In particular, the SV's reference point is defined by the center position of its front bumper and the POV's reference line is defined by a line passing through the center of its rear bumper and perpendicular to its heading direction.

\textbf{Time Headway (THW)}~\cite{Jansson2005CollisionAT} is similar to GT, except that the POV's reference line is defined w.r.t. its front bumper instead of the rear bumper.

\textbf{Level of Unsafety (LU)}~\cite{barcelo2003safety} measures the risk involved in a potential rear-end collision, expressed through a combination of the SV speed, the relative POV speed w.r.t. to the SV, and the POV instantaneous deceleration compared to the maximum allowed deceleration value.

\textbf{Collision Rate (CR)} defines the number of collisions per unit distance covered by the SV. Although a clear source of the initial proposal of CR is not available in the literature, it remains as an intuitively effective and commonly adopted performance measure of vehicle safety.

\textbf{Failure-free Miles Risk Inference (FMRI)}~\cite{kalra2016driving} infers the reliability $R$ (or fatality rate $1-R$) with a certain confidence level based on the number of failure-free miles driven. The original paper~\cite{kalra2016driving} seeks to illustrate the significantly large number of on-road safe driving miles required to demonstrate ADS's safety and reliability, therefore encouraging alternate methods of testing.  

\textbf{$\bar{\epsilon}\alpha$-Almost Safe Set}~\cite{weng2023finite} characterizes the SV's safety performance through a set of states $\Phi \subseteq O_l$ (approximated as an $\alpha$-shape) and the provably unbiased expected probability ($1-\bar{\epsilon}$) for the SV to remain inside that set. The metric also includes other set characterizations that are indicative of the vehicle's safety performance revealed through the input data set, such as the density ($|\Phi|/|\mathcal{D}_{O_l}|$) and the occupancy ($|\Phi|/|O_l|$). Although $\bar{\epsilon}\alpha$-ASS is the only metric empirically studied in this paper taking the notion of a ``set" as the output, set propagation has wide applications in safety testing with many other metric variants~\cite{althoff2021set} as we have discussed in Section~\ref{sec:oliterature}. 

\section*{Acknowledgment}
This work was supported by National Highway Traffic Safety Administration (NHTSA), U.S. Department of Transportation. This is the work of a government and is not subject to copyright protection. Foreign copyrights may apply. The government under which this work was written assumes no liability or responsibility for the contents of this work or the use of this work, nor is it endorsing any manufacturers, products, or services cited herein and any trade name that may appear in the work has been included only because it has been deemed essential to the contents of the work. Positions and opinions advanced in this work are those of the authors and not necessarily those of NHTSA. Responsibility for the content of the work lies solely with the authors. 

\bibliographystyle{IEEEtran}
\bibliography{output}

\end{document}